\documentclass[10pt,journal,compsoc]{IEEEtran}
\makeatletter
\def\endthebibliography{%
	\def\@noitemerr{\@latex@warning{Empty `thebibliography' environment}}%
	\endlist
}

\makeatother

\usepackage[nocompress]{cite}
\usepackage{stfloats}
\usepackage{epsfig}
\usepackage{graphicx}
\usepackage{amsmath}
\usepackage{amssymb}
\usepackage{amsfonts}
\usepackage{multirow}
\usepackage{url}
\usepackage{threeparttable}
\usepackage{booktabs}
\usepackage{dcolumn}
\usepackage{float} 
\usepackage{subfigure}
\usepackage{enumitem}
\usepackage{float}
\usepackage{lipsum}
\usepackage{overpic}

\usepackage{ragged2e}
\usepackage[table, dvipsnames]{xcolor}

\usepackage{soul}
\usepackage{color}
\usepackage{colortbl}

\newcolumntype{d}[1]{D{.}{.}{#1}}

\newlength\savewidth\newcommand\shline{\noalign{\global\savewidth\arrayrulewidth
  \global\arrayrulewidth 1pt}\hline\noalign{\global\arrayrulewidth\savewidth}}
\newcommand{\tablestyle}[2]{\setlength{\tabcolsep}{#1}\renewcommand{\arraystretch}{#2}\centering\footnotesize}

\def\up#1{\textcolor[rgb]{0,0.75,0.25}{$\uparrow${#1}}}

\definecolor{water}{RGB}{0, 0, 255}
\definecolor{non-vegetated-ground}{RGB}{128, 128, 128}
\definecolor{low-vegetation}{RGB}{0, 128, 0}
\definecolor{tree}{RGB}{0, 255, 0}
\definecolor{buildings}{RGB}{128, 0, 0}
\definecolor{playgrounds}{RGB}{255, 0, 0}

\definecolor{non_damage}{RGB}{0, 237, 253}
\definecolor{minor_damage}{RGB}{0, 255, 0}
\definecolor{major_damage}{RGB}{237, 142, 0}
\definecolor{destroyed}{RGB}{255, 0, 0}

\newcommand{\mpara}[2]{\vspace{#1}\noindent\textbf{#2}}

\newcommand{\best}[1]{\textbf{#1}}

\newcommand{\I}{\mathbf{I}}

\newcommand{\Sx}[1]{\mathbf{S}_{#1}}

\definecolor{create}{RGB}{46, 117, 182}
\definecolor{remove}{RGB}{197, 90, 17}
\definecolor{edit}{RGB}{95, 128, 64}

\definecolor{mod}{HTML}{f8e5d8}

\newcommand{\Create}{\textcolor{create}{\bf C}}
\newcommand{\Remove}{\textcolor{remove}{\bf R}}

\newcommand{\gr}{\rowcolor[gray]{.95}}

\definecolor{ftcolor}{HTML}{412F8A}
\definecolor{zscolor}{HTML}{fc8e62}

\newcommand{\ft}{\textcolor{ftcolor}{$\bullet$\,}}
\newcommand{\zs}{\textcolor{zscolor}{$\bullet$\,}}

\usepackage[pagebackref=false,breaklinks=true,letterpaper=true,colorlinks,bookmarks=false,citecolor={blue},linkcolor={blue}]{hyperref}

\usepackage{caption}
\usepackage[margin=4pt,font=footnotesize,labelfont=bf,labelsep=endash,tableposition=top]{caption}

\soulregister\citep7
\soulregister\cite7
\soulregister\ref7
\soulregister\subsection7

\definecolor{regular}{rgb}{1,1,1}
\definecolor{revision}{rgb}{1,0.92,0.24}
\sethlcolor{revision}

\begin{document}

\title{Changen2: Multi-Temporal Remote Sensing Generative Change Foundation Model
}
\author{Zhuo Zheng, Stefano Ermon, Dongjun Kim, Liangpei Zhang, Yanfei Zhong
\IEEEcompsocitemizethanks
{
\IEEEcompsocthanksitem Z. Zheng, S. Ermon, and D. Kim are with the Department of Computer Science, Stanford University (zhuozheng@cs.stanford.edu, ermon@cs.stanford.edu, dongjun@stanford.edu).
\IEEEcompsocthanksitem Y. Zhong and L. Zhang are with the State Key Laboratory of Information Engineering in Surveying, Mapping and Remote Sensing, Wuhan University, Wuhan, Hubei 430072, China.
\IEEEcompsocthanksitem Corresponding author: Stefano Ermon, Yanfei Zhong
\protect\\
E-mail: ermon@cs.stanford.edu, zhongyanfei@whu.edu.cn
\protect \\
\IEEEcompsocthanksitem A preliminary version of this work has been presented in ICCV
2023 \cite{changen}.
}
}%

\IEEEtitleabstractindextext{\begin{abstract}\justifying
Our understanding of the temporal dynamics of the Earth's surface has been significantly advanced by deep vision models, which often require a massive amount of labeled multi-temporal images for training.
However, collecting, preprocessing, and annotating multi-temporal remote sensing images at scale is non-trivial since it is expensive and knowledge-intensive.
In this paper, we present scalable multi-temporal change data generators based on generative models, which are cheap and automatic, alleviating these data problems. 
Our main idea is to simulate a stochastic change process over time.
We describe the stochastic change process as a probabilistic graphical model, namely the generative probabilistic change model (GPCM), which factorizes the complex simulation problem into two more tractable sub-problems, i.e., condition-level change event simulation and image-level semantic change synthesis. 
To solve these two problems, we present Changen2, a GPCM implemented with a resolution-scalable diffusion transformer which can generate time series of remote sensing images and corresponding semantic and change labels from labeled and even unlabeled single-temporal images.
Changen2 is a ``generative change foundation model'' that can be trained at scale via self-supervision, and is capable of producing change supervisory signals from unlabeled single-temporal images.
Unlike existing ``foundation models'', our generative change foundation model synthesizes change data to train task-specific foundation models for change detection.
The resulting model possesses inherent zero-shot change detection capabilities and excellent transferability. 
Comprehensive experiments suggest Changen2 has superior spatiotemporal scalability in data generation, e.g., Changen2 model trained on 256$^2$ pixel single-temporal images can yield time series of any length and resolutions of 1,024$^2$ pixels.
Changen2 pre-trained models exhibit superior zero-shot performance (narrowing the
performance gap to 3\% on LEVIR-CD and approximately 10\% on both S2Looking and SECOND, compared to fully supervised counterpart) and transferability across multiple types of change tasks, including ordinary and off-nadir building change, land-use/land-cover change, and disaster assessment. 
\end{abstract}

}

\maketitle

\IEEEdisplaynontitleabstractindextext

\IEEEpeerreviewmaketitle

\section{Introduction}
\IEEEPARstart{C}{hange} detection is one of the most fundamental Earth vision tasks, where the goal is to understand the temporal dynamics of the Earth's surface.
Tremendous progress in change detection has been achieved by joint efforts of the remote sensing and computer vision communities.
Deep change detection models \cite{daudt2018fully, changestar, changeos, zheng2022changemask} based on Siamese networks \cite{siamese} have dominated in recent years.
The key to their success lies in large-scale labeled training datasets \cite{gupta2019xbd, shen2021s2looking, DEN, tian2022large}.
However, building a large-scale remote sensing change detection datasets is difficult and expensive
because collecting (identifying image series where change events occur), preprocessing (e.g., the need for extra image registration), and annotating remote sensing image time series requires more expertise and effort compared with preparing a dataset for single-image tasks.

Synthetic data, as an alternative, is a promising direction to alleviate labeled data requirements.
Graphics-based \cite{bourdis2011constrained, kolos2019procedural, song2024syntheworld} and data augmentation-based \cite{copypaste} approaches are currently the two main approaches for change data synthesis in the remote sensing domain.
Graphics-based methods synthesize images by rendering manually constructed 3D models, thus providing flexible control abilities for geometric and radiance features, e.g., viewpoint, azimuth, and sunlight.
Data augmentation-based methods, which require no graphic expertise, create new image pairs by inserting object instances into existing bitemporal image pairs \cite{chen2021adversarial}.
Despite their potential, the scale and diversity of conventional synthetic change datasets are limited.
Graphics-based approaches are hindered by the labor-intensive nature of 3D modeling, whereas data augmentation-based approaches are limited by the scale of existing change datasets.
Additionally, while synthetic change data primarily enhances the performance of change detection models on real-world data, the relationship between the properties (e.g., quality and diversity) of synthetic data and the transferability of the features learned from synthetic data remains unclear.

In this paper, we present new scalable multi-temporal change data generators based on generative change modeling.
Our data generator aims to generate realistic and diverse multi-temporal images and dense labels from a single-temporal image and its annotations (e.g., semantic segmentation mask, object contour), by \textit{simulating the change process}.
To this end, we first describe the stochastic change process as a probabilistic graphical model, namely generative probabilistic change model (GPCM), considering pre-event and post-event images $\I_t$ and $\I_{t+1}$ along with their conditions $\Sx{t}$ and $\Sx{t+1}$ (e.g., semantic mask) as random variables, as shown in Fig.~\ref{fig:gpcm}. 
The change is always driven by the event, i.e., post-event data depends on pre-event data.
Meanwhile, we assume this simulation is a generative task, that is, the image depends on its condition.
Based on these two conditions, the whole problem can be simplified into two subproblems, i.e., the change event simulation at the condition (semantic label) level and the semantic change synthesis at the image level.

To solve the above two subproblems, we propose a generative model called \textit{Changen2}, which is a GPCM implemented with diffusion models (DMs).
Our Changen2 creates objects, removes objects, or edits object's attributes in the semantic mask at time $t$, as a stochastic change event, to generate a new semantic mask at time $t+1$, and synthesizes a post-event image by progressively applying simulated semantic changes to the pre-event image. 
Changen2 can be trained on single-temporal images using their semantic mask labels as conditions.
To enable Changen2 to learn from large-scale unlabeled earth observation data, we further design a self-supervised learning approach to train Changen2 with unlabeled single-temporal images at scale.
Our self-supervision at scale comes from object contours generated by segment anything model (SAM) \cite{sam}.
We use object contours as the condition to train the diffusion model of Changen2 and simulate change events by removing objects.
The post-event image can then be synthesized by an unlabeled pre-event image and post-event object contours.
The change mask for this event is a binary mask indicating those areas of removed objects.
In this way, self-supervised Changen2 can synthesize change-labeled multi-temporal images from unlabeled single-temporal images, yielding change supervision at scale.
We refer to this model, capable of generating change supervision at scale in a self-supervised generative way as the ``generative change foundation model'' to distinguish the concepts of ``generative foundation model'', e.g., DiffusionSat \cite{diffusionsat}, and other MAE-based foundation models, e.g., SatMAE \cite{satmae}.

As a demonstration, we use Changen2 to generate three large-scale synthetic change detection datasets, including a building change detection dataset (\texttt{Changen2-S1-15k}) with diverse object properties (e.g., scale, shape, position, orientation) and two change types, a semantic change detection dataset (\texttt{Changen2-S9-27k}) with up to 38 change types, and a class-agnostic change detection dataset (\texttt{Changen2-S0-1.2M}).
The change detector pre-trained on these synthetic datasets has superior transferability on real-world change detection datasets, outperforming state-of-the-art SA-1B \cite{sam} and Satlas \cite{satlas} pre-training.
Additionally, it exhibits outstanding zero-shot prediction capability, significantly improving over AnyChange \cite{anychange} by 16.4\% F$_1$ at the pixel level SECOND benchmark.
This narrows the performance gap to 3\% on LEVIR-CD, and approximately 10\% on both S2Looking and SECOND, compared to fully supervised counterparts.
More importantly, based on our Changen2, we find that the temporal diversity of synthetic change data is a key factor in ensuring transferability after model pre-training.
This is the key to the success of synthetic change data pre-training.

The main contributions of this paper are summarized as follows:
\begin{itemize}[leftmargin=*]
\item \textbf{Generative change modeling} decouples the complex stochastic change process simulation to more tractable change event simulation and semantic change synthesis.
\item \textbf{Generative change foundation model}, Changen2, employing a novel resolution-scalable diffusion transformer architecture, can generate time series of remote sensing images and corresponding semantic and change labels from single-temporal images.
Our model can be trained at scale using both labeled data using supervision and unlabeled data using self-supervision.
This model provides a new approach for producing task-tailored remote sensing foundation models.
\item \textbf{Globally distributed synthetic change datasets}.
Based on three globally distributed single-temporal satellite image datasets, we use Changen2 to generate three globally distributed change detection datasets: $\texttt{Changen2-S1-15k}$ (a building change dataset with 15k pairs and 2 change types), $\texttt{Changen2-S9-27k}$ (an urban land-use/land-cover change dataset with 27k pairs and 38 change types), and $\texttt{Changen2-S0-1.2M}$ (a class-agnostic change dataset with 1.2 million pairs and innumerable change types).
Our synthetic change data pre-training empowers change detection models with better transferability and zero-shot prediction capability, comparable to fully supervised counterparts in ordinary scenarios.
$\texttt{Changen2-S0-1.2M}$ pre-trained foundation model significantly outperforms seven state-of-the-art self-supervised remote sensing foundation models.
Additionally, for the first time, our foundation model bridges the performance gap with two strong supervised foundation models (Satlas \cite{satlas} and SAM \cite{sam}).
\end{itemize}

This work extends our conference paper \cite{changen} and substantially improves upon it in three key aspects:

\mpara{0mm}{1)} Our previous version, Changen, is a GPCM implemented with GAN, focusing on the generation of single-class object changes.
To improve generation capability and enable more general multi-class object change generation, we propose Changen2, a GPCM implemented with our latent diffusion model, the resolution-scalable diffusion transformer (RS-DiT), to improve the quality and correspondence between multi-temporal images and their change labels in synthetic change data.
We also present a new change event simulation, i.e., attribute editing, to better support multi-class change synthesis.

\mpara{0mm}{2)} To enable our model to learn the distribution of massive, unlabeled remote sensing images, we propose a self-supervised learning algorithm to train Changen2. 
This approach eliminates the need for manual annotations and allows Changen2 to leverage vast amounts of unlabeled Earth observation data.

\mpara{0mm}{3)} We build three globally distributed synthetic change datasets, including up to 38 change types and over one million image pairs.
The model pre-trained on these synthetic change data exhibits superior transferability and more importantly possesses zero-shot change detection capability, which existing remote sensing foundation models cannot achieve.
Besides, these datasets contribute to the development of remote sensing synthetic data and support subsequent algorithm research, including transfer learning from synthetic to real-world scenarios, more advanced synthetic data pre-training strategies, and zero-shot change detection technologies.

\section{Related Work}
\label{sec:related_work}

\mpara{0em}{Change Data Synthesis.}
Based on the computer graphics, the early studies mainly used game engines to generate realistic remote sensing images from existing assets.
For example, the AICD dataset~\cite{bourdis2011constrained} consists of 1,000 pairs of 800$\times$600 images that are automatically generated by the built-in rendering engine of a computer game.
However, this dataset has low diversity and graphics quality due to limited assets and an underdeveloped rendering engine.
To improve the diversity and realism of synthesized images, a semi-automatic data generation pipeline \cite{kolos2019procedural} is proposed to synthesize a change detection dataset, which adopts cartographic data for manual 3D scene modeling and a professional game engine for rendering.
After we propose to use generative models to synthesize changes \cite{changen},
SyntheWorld \cite{song2024syntheworld} integrates procedural 3D modeling techniques with generative models \cite{achiam2023gpt,rombach2022high}, yielding more realistic land cover and building change datasets.
Another way is based on data augmentation, especially Copy-Paste \cite{copypaste}, e.g., IAug \cite{chen2021adversarial} adopts SPADE \cite{spade} to generate object instances and randomly pastes them into an existing change detection dataset to increase the number of training samples.
Our work introduces a new perspective, i.e., generative change modeling, which exclusively uses generative models to synthesize change directly.
Benefiting from a deep generative model, our dataset generator is automatic and relies only on single-temporal data, ensuring the scalability and reducing the cost of synthetic data generation.

\mpara{0em}{Semantic Image Synthesis.}
Recent studies \cite{pix2pixHD, spade, zhu2020sean, tan2021diverse, shi2022retrieval, oasis} mainly focus on translating the semantic segmentation label to the image based on different conditional GANs \cite{cGAN}.
pix2pix \cite{pix2pix} and pix2pixHD \cite{pix2pixHD}, as early representative methods, directly use the semantic mask to generate the image.
SPADE \cite{spade} reveals that normalization layers tend to ``wash away'' semantic information, and propose a spatially adaptive conditional normalization layer to incorporate semantic information and spatial layout.
The following studies further improve SPADE on region-level control ability \cite{zhu2020sean}, diversity \cite{tan2021diverse}, better trade-off between fidelity and diversity \cite{shi2022retrieval}.
These methods strongly depend on VGG-based perceptual loss \cite{dosovitskiy2016generating, johnson2016perceptual} to guarantee image quality; however, this increases the complexity of the whole pipeline.
To simplify GAN models for semantic image synthesis, OASIS \cite{oasis} reveals that the segmentation-based discriminator is the key to allowing the GAN model to synthesize high-quality images with only adversarial supervision.
Diffusion \cite{ddpm,sohl2015deep} and score-based generative models \cite{hyvarinen2005estimation,song2019generative,song2021scorebased}, as another widely used generative model, have achieved impressive results in image synthesis.
Latent diffusion models \cite{rombach2022high}, as one of the representative diffusion models, have been widely used in conditional image generation due to their efficiency and effectiveness.
Extra condition control \cite{controlnet} and transformer architectures \cite{uvit, dit} make latent diffusion models exhibit outstanding performance in semantic image synthesis.
Our work extends conventional semantic image synthesis in temporal dimensionality, enabling semantic time series synthesis from single-temporal data with better spatiotemporal scalability and consistency.

\mpara{0em}{Remote Sensing Visual Foundation Models.}
Our synthetic data includes images and labels, mainly used for model pre-training, namely synthetic data pre-training.
Current remote sensing self-supervised pre-training approaches are mainly based on contrastive learning \cite{moco, mocov2}, masked image modeling \cite{beit,simmim,mae}, or their combination.
Their pre-trained models are usually referred to as remote sensing foundation models (RSFMs).
Based on general-purpose MoCo \cite{moco}, GASSL \cite{gassl} introduces multi-temporal images to construct positive pairs and geo-location-based loss to learn geography-aware representation.
SeCo \cite{seco} and CACo \cite{caco} adopt seasonal and change-aware contrast to learn transferable representation, respectively.

Masked autoencoders (MAE) \cite{mae} are confirmed to have better scalability.
Consequently, RSFMs have gradually shifted towards MAE-based approaches in recent years.
For example, SatMAE \cite{satmae} design data-specific masking strategies to make MAE learn from multi-temporal and multi-spectral images.
SpectralGPT \cite{hong2024spectralgpt} improves the masking strategy tailored for multi-spectral images via 3D masking \cite{feichtenhofer2022masked, tong2022videomae} and spectral reconstruction.
Scale-MAE \cite{scalemae} and cross-scale MAE \cite{csmae} improve MAE with scale-aware representations through positional encoding and cross-scale consistency, respectively.
SatMAE++ \cite{satmaepp} improves SatMAE via multi-scale reconstruction.
Apart from those self-supervised pre-training methods, Satlas \cite{satlas} collects a large-scale multi-task labeled dataset to pre-train deep models via supervised learning, yielding promising improvements on multiple remote sensing image understanding tasks.

These RSFMs can provide a good starting point, however, the specific capabilities (e.g., classification, segmentation, change detection) require fine-tuning on task-specific labeled data to be achieved.
There is a large capability gap between existing RSFMs and those foundation models with zero-shot prediction capabilities, e.g., SAM \cite{sam}, CLIP \cite{clip}, and GPTs \cite{gpt2,gpt4}.
Our work provides a new feasible roadmap towards RSFM with zero-shot prediction capability, i.e., synthetic data pre-training.
Without any fine-tuning, our models pre-trained on synthetic change data have superior zero-shot change detection capability for unseen data distribution.

\section{Generative Probabilistic Change Model}
\label{sec:method}
\begin{figure}[ht]
\centering
\includegraphics[width=0.95\linewidth]{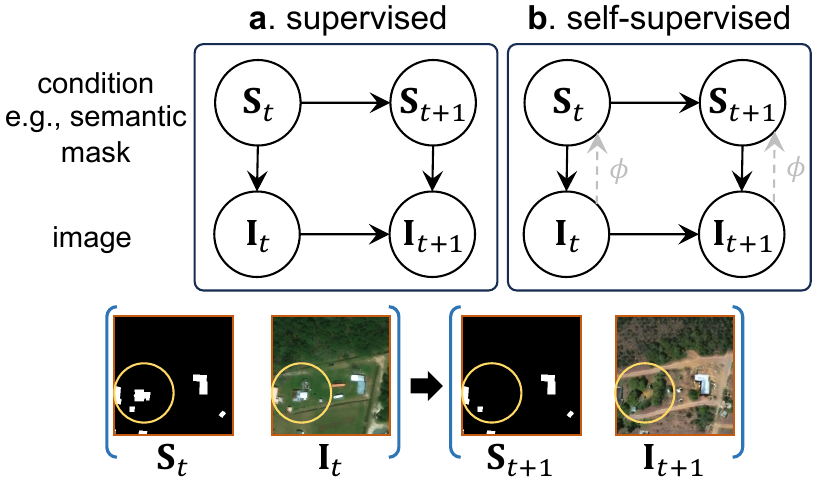}
\caption{\textbf{Generative Probabilistic Change Model} (GPCM).
The bottom subfigure is a case of semantic mask as the condition.
$\phi$ denotes an editable condition extractor to provide self-supervision.
}
\label{fig:gpcm}
\end{figure}

The main idea of our change data generation is to simulate the stochastic change process starting from each single-temporal image and its condition, e.g., semantic mask.
To this end, we frame the stochastic change process as a probabilistic graphical model shown in Fig.~\ref{fig:gpcm}, to describe the relationship between the distributions of variables (i.e., images, semantic masks) over time.
In this way, we can parameterize several smaller factors instead of directly parameterizing the high-dimensionality joint distribution over all variables, simplifying this simulation problem.
Based on this modeling framework, we provide implementations (Changen \cite{changen} and Changen2 in this work) with two types of deep generative models (GAN and DPM) to synthesize the multi-temporal change detection dataset from single-temporal data.

\subsection{Generative Change Modeling Framework}
The image $\I_t$ has predefined pixel-wise semantics $\Sx{t}$ at time $t$.
Given a stochastic change event occurred in the time period $t$ to $t+1$, the image $\I_{t}$ gradually evolves into the image $\I_{t+1}$ with a semantic transition from $\Sx{t}$ to $\Sx{t+1}$.
The joint distribution of this stochastic change process is denoted as $P_{\rm scp}:= P(\Sx{t+1}, \Sx{t}, \I_{t+1}, \I_{t})$.
Based on the chain rule, we factorize $P_{\rm scp}$ as $P(\Sx{t+1}, \I_{t+1}| \Sx{t}, \I_{t})P(\Sx{t}, \I_{t})$, where post-event data depends on pre-event data since the change is always event-driven (e.g., human activities, nature hazards).
We make an assumption that simulation is a generative task, that is, semantics to images, the graph structure of the set of these four random variables can be described as shown in Fig.~\ref{fig:gpcm}a.
The following factorization can be obtained:
\begin{equation}
   P_{\rm scp} = P(\I_{t+1}|\Sx{t+1}, \I_{t})P(\Sx{t+1}|\Sx{t})P(\I_{t}|\Sx{t})P(\Sx{t})
\end{equation}
where the semantics distribution $P(\Sx{t})$ and corresponding conditional image distribution $P(\I_{t}| \Sx{t})$ can be approximately seen as two known distributions because the single-temporal data is given for this simulation problem.
This means that the semantic transition distribution $P(\Sx{t+1}|\Sx{t})$ and the conditional image distribution $P(\I_{t+1}|\Sx{t+1}, \I_{t})$ need to be further estimated, to sample from $P_{\rm scp}$.
Through above modeling, the whole simulation problem can be decoupled to two subproblems, i.e., the \textit{change event simulation} to approximate $P(\Sx{t+1}|\Sx{t})$ and the \textit{semantic change synthesis} to approximate $P(\I_{t+1}|\Sx{t+1}, \I_{t})$.

\subsection{Change Event Simulation}
\begin{figure*}[htb]
\centering
\includegraphics[width=\linewidth]{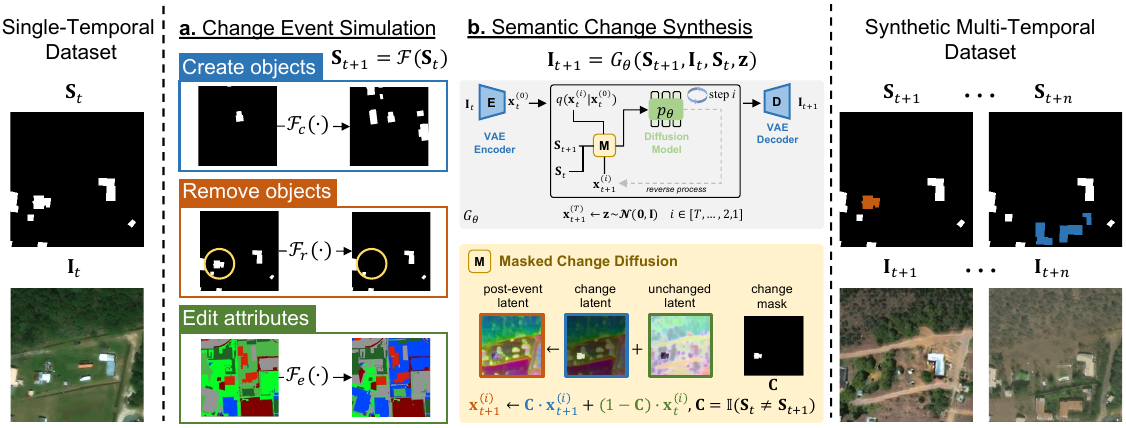}
\caption{\textbf{Our Changen2 framework}.
The change event simulation enables \textcolor{create}{adding}, \textcolor{remove}{removing} objects, and \textcolor{edit}{editing} attributes of objects in the semantic mask at time $t$ to customize new semantic masks at times $t+1~{\rm to}~n$.
For the semantic change synthesis, the new images at times $t+1~{\rm to}~n$ will be synthesized by iteratively conditional denoising on the image at time $t$.
Changen2 can generate the multi-temporal dataset with controllable scene layout, object property (e.g., scale, position, orientation, semantics, see $\I_{t+n}$), and change event.
Legend: \textcolor{create}{\textbf{Create}}; \textcolor{remove}{\textbf{Remove}}.
}
\label{fig:changen2}
\end{figure*}

Sampling from $P(\Sx{t+1}|\Sx{t})$ is to obtain $\Sx{t+1}$ as a conditional guidance for the subsequent semantic change synthesis.
We consider three common change events, i.e., the construction of new objects, the destruction of existing objects, and the change of object attributes in the real world.
To simulate these three events on the semantic level, we design three rule-based functions, as shown in Fig.~\ref{fig:changen2}a.

\mpara{0mm}{Object Creation} $\mathcal{F}_{\rm c}(\cdot): R^{h\times w}\rightarrow R^{h\times w}$.
We first uniformly sample some instances from the semantic masks.
The selected instances are pasted into the rest area of the semantic mask to simulate object creation.

\mpara{0mm}{Object Removal} $\mathcal{F}_{\rm r}(\cdot): R^{h\times w}\rightarrow R^{h\times w}$.
As with object creation, we first uniformly sample some instances from the semantic masks.
To remove these instances, we assign the region of selected instances with a background pixel value in the semantic mask.

\begin{figure}[t]
\centering
\includegraphics[width=\linewidth]{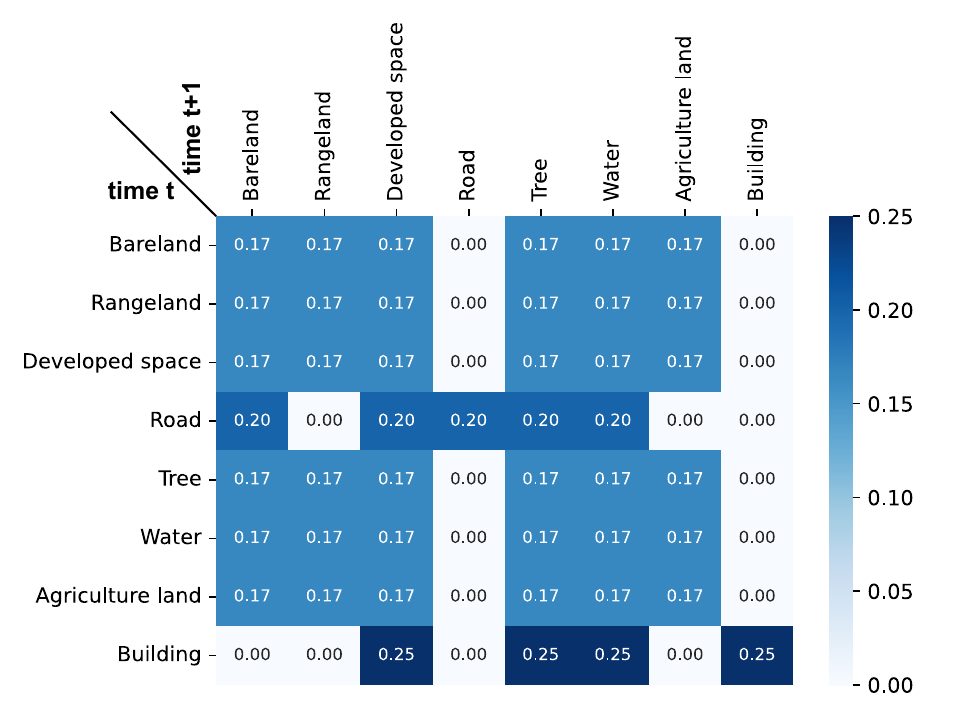}
\caption{\textbf{Attribute Edit: Customized Semantic Transition Matrix}.
We demonstrate a uniformly sampling case for the category system of the OpenEarthMap dataset.
Based on this semantic transition matrix, we can inject the change class prior into synthetic change data, thereby achieving the desired dataset as application scenarios require.
}
\label{fig:transition}
\end{figure}

\mpara{0mm}{Attribute Edit} $\mathcal{F}_{\rm e}(\cdot): R^{h\times w}\rightarrow R^{h\times w}$.
Unlike object creation and removal, editing attributes does not change the spatial layout of objects.
Only the object attribute will be changed.
For example, in terms of semantics, the object region is changed from bareland to water.
To simulate desired change events, we propose to define a semantic transition matrix as shown in Fig.~\ref{fig:transition}, which is a customized semantic transition matrix based on the category system of the OpenEarthMap dataset \cite{oem}.
By assigning transition probability, we can control the change class distribution of synthetic change data to meet various application scenarios.
This design makes our Changen2 capable of introducing change prior information as guidance.
By default, we assign transition probabilities via uniform distribution to build our generative change foundation models.
Each instance will adjust its semantic attribute based on the pre-defined semantic transition matrix, thus resulting in the next semantic mask over time, as shown in Fig.~\ref{fig:changen2}a.

\begin{figure}[htb]
\centering
\includegraphics[width=\linewidth]{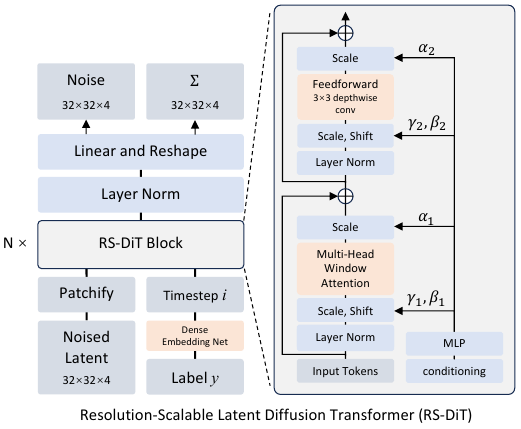}
\caption{\textbf{Network architecture of RS-DiT}.
Based on DiT architecture, we make two small but important improvements: (i) remove absolute position embedding and insert 3$\times$3 depthwise convolution in FFN; (ii) replace global self-attention with local window attention.
In addition, we introduce a dense embedding network to encode dense conditions, thereby enabling tasks that require dense conditional image generation.
Our improvements are \textcolor{Apricot}{\bf highlighted in color}.
}
\label{fig:rsdit}
\vspace{-4mm}
\end{figure}

\subsection{Changen: GANs as semantic change synthesizer}
We briefly recap the Changen \cite{changen} to provide necessary preliminary.
To solve the above two sub-problems, we present a parameterization for GPCM with the learning-free function family $\mathcal{F}(\cdot)$ for change event simulation and the well-designed deep generative model $G$ for semantic change synthesis, as follows:
\begin{align}
  & \Sx{t+1} = \mathcal{F}_{\{c, r, e\}}(\Sx{t})   \\
  & \I_{t+1}  = G_{\theta}(\Sx{t+1}, \I_{t}, \Sx{t}, \mathbf{z}), \mathbf{z}\sim \mathcal{N}(\mathbf{0}, \mathbf{1})
\end{align}
where $\Sx{t}$ is only used for non-parameterized computation. 

Changen is an instance of GAN-based GPCM.
The network architecture of Changen includes an image encoder that provides pre-event image guidance and a decoder that serves as the conditional generator, mapping noise to an image.
It is non-trivial to leverage pre-event image guidance since there is a trivial solution caused by ``feature leakage''.
This causes the network to merely replicate the pre-event image instead of generating new content given the condition.
To address ``feature leakage'', \textit{masked transition layer} is proposed in Changen, which compels the network to accurately generate post-event image based on the given conditions.
Due to the absence of the ground truth of post-event images, \textit{bitemporal adversarial learning} is proposed, which allows Changen to be trained using only pre-event images and their semantic masks.
We can use synthetic change data from Changen to integrally pre-train change detection models for zero-shot change detection on unseen data distribution and good starting points for subsequent fine-tuning.

Although Changen has exhibited a promising approach for change data generation, there are still limitations to its application in real-world scenarios.
We argue that its limitations primarily include insufficient zero-shot prediction performance (a significant performance gap compared to supervised models, e.g., on LEVIR-CD dataset, previous state-of-the-art fully supervised models achieve nearly 92\% F$_1$, while zero-shot models only reach 62.8\% F$_1$), lack of support for multi-class change generation, and requirement for single-image annotations.
These problems significantly limit the applications of synthetic change data and obscure its potential.
We resolve these three challenging problems via several novel technical improvements, described next.

\subsection{Changen2: DMs as semantic change synthesizer}
\mpara{0mm}{Motivation}.
The above limitations mainly point to the conditional generation capability and supervisory signals of Changen.
To this end, we propose Changen2 framework, which adopts a tailored transformer-based diffusion model to improve conditional generation capability.
Furthermore, we propose a self-supervised learning approach (see Sec.~\ref{sec:ssl}) to train Changen2 with unlabeled single images, thereby eliminating the requirement for manual single-image annotations.

\begin{figure}[htb]
\centering
\includegraphics[width=\linewidth]{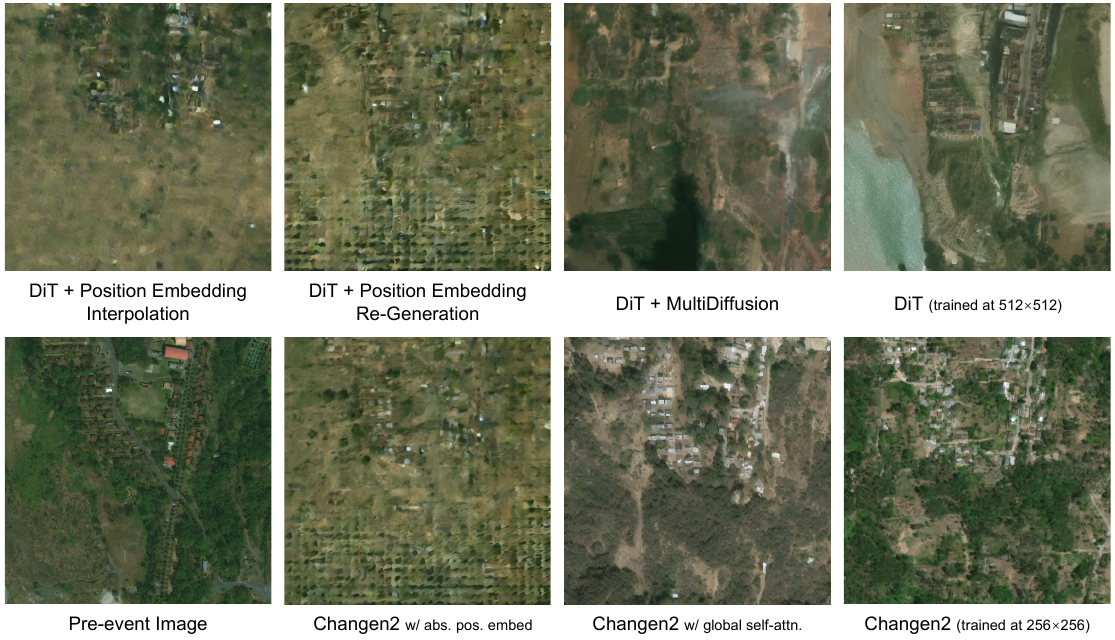}
\caption{\textbf{Analysis and Ablation: Spatial Resolution Scalability (256$^2$ px to 512$^2$ px)}.
All generated 512$\times$512 images are synthesized post-event images \textit{without pre-event image guidance}.
The top left three results are generated from a DiT model trained at 256$\times$256.
}
\label{fig:256_to_512}
\vspace{-4mm}
\end{figure}

\mpara{1mm}{Network Architecture: RS-DiT}.
Changen2 aims to generate high-resolution time series from a single image, following the spatiotemporal nature of earth observation.
However, training a contemporary transformer-based diffusion model on high-resolution images is extremely computationally expensive \cite{multidiffusion, kim2024pagoda}.
Low-resolution (e.g. 256$^2$) trained model (e.g., diffusion transformer (DiT) \cite{dit}) struggle to generate high-resolution images (e.g., 512$^2$ and 1,024$^2$).

To address this resolution scalability problem, we propose a resolution-scalable diffusion transformer (RS-DiT) as the conditional noise prediction network $p_\theta$ of Changen2, as shown in Fig.~\ref{fig:rsdit}.
Based on the DiT architecture, we identify a key factor that constrain resolution scalability: absolute positional embeddings.
Specifically, in terms of architecture, absolute positional embedding is the only barrier that restricts the input image resolution, and interpolating or regenerating these frequency-based positional embeddings for a new resolution is ineffective (Fig.~\ref{fig:256_to_512}, top-left two).
Besides, the quadratic computation complexity of a global self-attention layer leads to a huge cost for high-resolution image generation.
To address these problems, we introduce two small but important modifications to DiT architecture, thus making DiT resolution scalable: (i) remove absolute positional embeddings and insert a 3$\times$3 depthwise convolution layer into the feed-forward network (FFN) of each DiT block, which provides relative position information via zero-padding \cite{islam2020much}; (ii) use local window attention layer in place of global self-attention layer in all DiT blocks except $4*i$-th block, e.g., $i = 1,2,3$ for DiT-B.
To incorporate dense labels as conditional input, we design a dense embedding network composed of 8 conv-layernorm-SiLU blocks, with 2$\times$ downsampling occurring between every two blocks.
The 2$\times$ downsampling is implemented by a 3$\times$3 convolution layer with a stride of 2.
This dense embedding network results in a total of 8$\times$ downsampling to match the input latent size.

\mpara{1mm}{Training Objective}.
Our model is a latent diffusion model, which is trained in latent space.
We use a VAE encoder to convert the image into its latent.
Given a single-temporal image latent $\mathbf{x}^{(0)}$, timestep $i$ (here allow us to use $i$ to denote diffusion timestep since $t$ denotes satellite imaging time), the diffusion forward process first perturb $\mathbf{x}^{(0)}$ by gradually adding Gaussian noise to the scaled input, obtaining $\mathbf{x}^{(i)}\sim q(\mathbf{x}^{(i)}|\mathbf{x}^{(0)}) = \mathcal{N}(\mathbf{x}^{(i)}; \sqrt{\Bar{\alpha}^{(i)}}\mathbf{x}^{(0)}, (1 - \Bar{\alpha}^{(i)})\mathbf{I})$, where $\Bar{\alpha}^{(i)}$ are hyperparameters dependent on a specific noise schedule.
Given semantic mask $\mathbf{c}$ as a condition, our model learns the reverse process $p_\theta(\mathbf{x}^{(i-1)}|\mathbf{x}^{(i)}, \mathbf{c})=\mathcal{N}(\mathbf{x}^{(i-1)}; \mu_\theta(\mathbf{x}^{(i)}, \mathbf{c}), \Sigma_\theta(\mathbf{x}^{(i)}, \mathbf{c}))$ by optimizing the variational lower bound \cite{kingma2013auto} of log-likelihood of $\mathbf{x}^{(0)}$.
Following DiT's training strategy \cite{dit, nichol2021improved}, we train the covariance $\Sigma_\theta(\mathbf{x}^{(i)}, \mathbf{c})$ via full variational lower bound and train the mean $\mu_\theta(\mathbf{x}^{(i)}, \mathbf{c})$ reparameterized as noise prediction $\epsilon_\theta(\mathbf{x}^{(i)}, \mathbf{c})$ with a simplified objective (Eq.~\ref{eq:sim}):
\begin{equation}\label{eq:sim}
    L(\theta) :=  \mathbb{E}_{\mathbf{x}^{(0)}, i, \mathbf{c}, \epsilon} ||\epsilon - \epsilon_\theta(\mathbf{x}^{(i)}, \mathbf{c})||_2^2,
\end{equation}
where $\epsilon\in\mathcal{N}(\mathbf{0},\mathbf{I})$.
This objective naturally involves only single-temporal images and their conditions, thus fundamentally avoiding the feature leakage problem in Changen, where the network uses the pre-event image to generate the post-event image.

\mpara{1mm}{Inference: Masked Change Diffusion}.
Building on a well-trained Changen2 model $p_\theta$ with single-temporal images, we further propose an iterative inference algorithm, i.e., masked change diffusion (Fig.~\ref{fig:changen2}b) for semantic change synthesis, leveraging the ``masked diffusion'' nature of diffusion models \cite{lugmayr2022repaint,controlnet}.
Given a pre-event image latent $\mathbf{x}_t^{(0)}$ and its semantic mask $\Sx{t}$, we first sample a post-event semantic mask $\Sx{t+1}$ via our change event simulation and then compute their change mask $\mathbf{C}=\mathbb{I}(\Sx{t}\neq\Sx{t+1})$, where $\mathbb{I}$ is an indicator function.
We initialize the post-event image latent $\mathbf{x}_{t+1}^{(T)}\sim \mathcal{N}(\mathbf{0}, \mathbf{I})$, where $T$ is the number of iterations.
To control temporal coherence and diversity, we have a pre-event image guidance ratio $\lambda\in[0,1]$, where only first $\lfloor\lambda T\rfloor$ steps consider this guidance.
For each timestep $i$ that requires pre-event image guidance, we perturb the pre-event image latent $\mathbf{x}_t^{(i)}\sim q(\mathbf{x}_t^{(i)}|\mathbf{x}_t^{(0)})$ via diffusion forward process and then compute a temporal-mixed latent based on the change mask, following \cite{lugmayr2022repaint}:
\begin{equation}
\mathbf{x}_{t+1}^{(i)}\leftarrow\mathbf{C}\cdot\mathbf{x}_{t+1}^{(i)} + (1-\mathbf{C})\cdot\mathbf{x}_{t}^{(i)}
\end{equation}
where the resulting latent preserves pre-event unchanged content and incorporates post-event changed content, thereby accurately corresponding to the change mask.
Next, we can compute the latent $\mathbf{x}_{t+1}^{(i-1)}$ at the next timestep $i-1$ using our Changen2 model:
\begin{equation}
p_\theta(\mathbf{x}_{t+1}^{(i-1)}|\mathbf{x}_{t+1}^{(i)}, \mathbf{c}=\Sx{t+1})
\end{equation}
where the condition $\mathbf{c}$ uses the post-event semantic mask $\Sx{t+1}$, ensuring the generated post-event image is semantically accurate.
By default, we adopt DDIM \cite{ddim} with $T=50$ to accelerate sampling.

\subsection{Self-Supervised Learning for Changen2}\label{sec:ssl}
Training a Changen2 model requires single-temporal images and their dense annotations as conditions.
However, the high cost and untimely acquisition of labels constrain Changen2's utilization of large-scale unlabeled Earth observation data.
To overcome this limitation, we propose a self-supervised learning approach to train Changen2 with large-scale unlabeled single-temporal images.
Our main idea is to exploit an editable condition as self-supervision, where this condition can be obtained from a single-temporal image itself.
As shown in Fig.~\ref{fig:gpcm}b, given an extractor $\phi$ for editable conditions, based on our GPCM, we can reformulate our generative model as follows:
\begin{equation}\label{eq:ssl_g}
\mathbf{I}_{t+1} = G_\theta(\Sx{t+1}{\rm=}\mathcal{F}(\phi(\mathbf{I}_t)), \mathbf{I}_t, \Sx{t}{\rm=}\phi(\mathbf{I}_t), \mathbf{z})
\end{equation}
where the editability of the condition is used to adapt the change event simulator $\mathcal{F}$, thereby constructing change labels.
In this way, this new generative model (Eq.~\ref{eq:ssl_g}) only relies on unlabeled single-temporal images and Gaussian noise, which is the foundation for self-supervised learning.

\begin{figure}[htb]
\centering
\includegraphics[width=\linewidth]{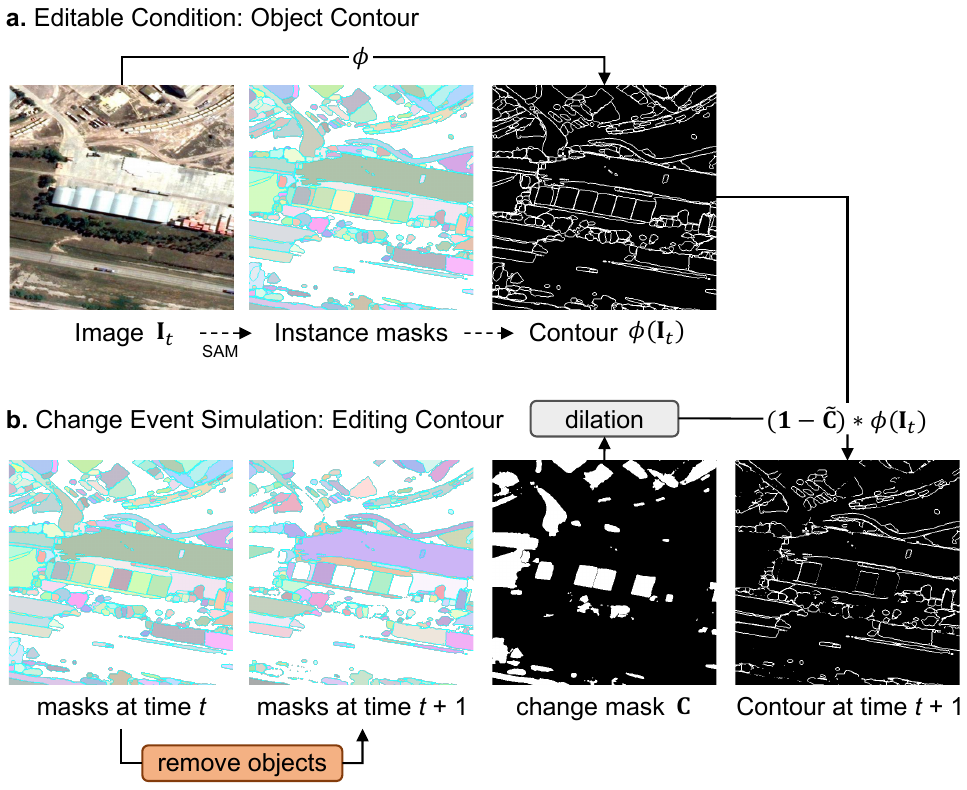}
\caption{\textbf{Contour-based implementation}. \textbf{a.} $\phi$: Object contour extractor and \textbf{b.} $\mathcal{F}$: Contour-based change event simulation.
}
\label{fig:ssl}
\vspace{-2mm}
\end{figure}

\begin{figure}[htb]
\centering
\includegraphics[width=0.9\linewidth]{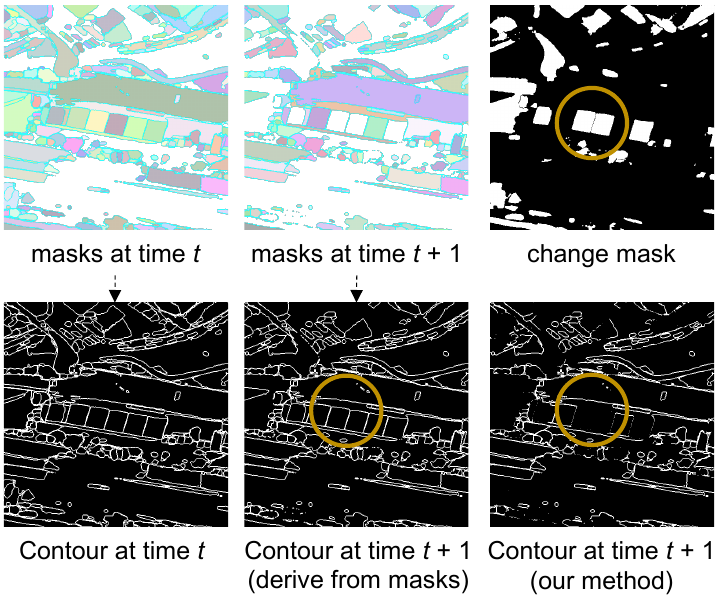}
\caption{\textbf{Next-time contour generation}.
The next-time contour map directly derived from masks corresponds wrong with the change mask since adjacent objects provide boundaries. 
Our change mask-based contour generation can correct this correspondence error.
}
\label{fig:correspondence}
\vspace{-2mm}
\end{figure}

\mpara{0mm}{Training}. Following the above formulation, we have a self-supervised training objective in place of Eq.~\ref{eq:sim} to train Changen2 models using the condition $\mathbf{c}{\rm=}\phi(\mathbf{I}_t)$:
\begin{equation}\label{eq:ssl}
L_{\rm ssl}(\theta) :=  \mathbb{E}_{\mathbf{x}_t^{(0)}, i, \mathbf{c}, \epsilon} ||\epsilon - \epsilon_\theta(\mathbf{x}_t^{(i)}, \mathbf{c}{\rm=}\phi(\mathbf{I}_t))||_2^2
\end{equation}
where $\mathbf{x}_t^{(0)}$ denotes the latent representation of image $\I_t$.
We train the covariance term $\Sigma_\theta(\mathbf{x}^{(i)}, \mathbf{c}{\rm=}\phi(\mathbf{I}_t))$ with the full variational lower bound.

\begin{figure}[htb]
\centering
\includegraphics[width=0.9\linewidth]{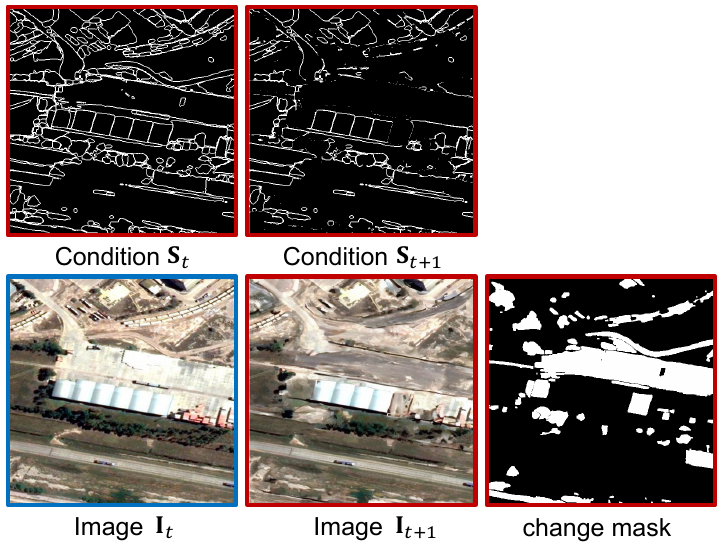}
\caption{\textbf{Self-Supervised Change Data Synthesis}.
Given a self-supervised Changen2 and an unlabeled single-temporal image $\I_t$ (\textcolor{NavyBlue}{blue box}), we can generate a next-time image $\I_{t+1}$ and a class-agnostic object change mask (\textcolor{Maroon}{red box}), thereby forming a training sample for change representation learning.
}
\label{fig:ssl_case}
\end{figure}

\mpara{0mm}{Inference}. Without any modification, we can still use masked change diffusion with a self-supervised Changen2 for semantic change synthesis.
As shown in Fig.~\ref{fig:ssl_case}, given an unlabeled single-temporal image $\I_t$, we first extract an object contour map $\phi(\I_t)$ as the condition $\Sx{t}$.
Through the change event simulation $\mathcal{F}$, we further obtain the next-time condition $\Sx{t+1} = \mathcal{F}(\phi(\I_t))$ and corresponding change mask $\mathbf{c}$.
Based on Eq.~\ref{eq:ssl_g}, the next-time image $\I_{t+1}$ can be generated.
The triplet $(\I_t, \I_{t+1}, \mathbf{c})$ is a training sample used to pre-train change detection models.

\mpara{1mm}{Editable Condition: Object Contour}.
As a demonstration, we propose an effective condition-editable function (Fig.~\ref{fig:ssl}a) that leverages SAM to extract object instance masks from an unlabeled single-temporal image $\mathbf{I}_t$ and then compute contours from these instance masks as the condition.
Its editability comes from the fact that the contour is derived from its instance, allowing us to edit the instance to modify the corresponding contour.

\mpara{1mm}{Change Event Simulation: Editing Contour}.
As shown in Fig.~\ref{fig:ssl}b, we further present a change event simulation method to produce the object contour at the next time $t+1$.
Given that masks generated by SAM are typically densely distributed due to dense point prompts, newly created objects can easily overlap with other objects, resulting in an unreasonable spatial layout.
Therefore, we only choose to remove objects as the change event.
By randomly removing objects at the time $t$, we can obtain object masks at the time $t+1$ and compute their binary object change mask $\mathbf{C}$.
To effectively remove corresponding contours, we first dilate the change mask and then compute the contour at the time $t+1$ using the dilated change mask $\widetilde{\mathbf{C}}$ to erase the contour $\phi(\I_t)$, i.e., $(1 - \widetilde{\mathbf{C}})*\phi(\I_t)$.
This design is non-trivial since directly computing contour from the object masks at the time $t+1$ cannot obtain accurate correspondence between the object change mask and bitemporal contours, as illustrated in Fig.~\ref{fig:correspondence}.
This is because the contours of the removed objects will be recovered from the contours of adjacent objects when generating contours directly from object masks.
This is why we incorporate the object change mask into contour generation.
Based on this indirect contour generation, we can obtain accurate correspondence between the object change mask and bitemporal contours.

\section{Experiments}
\label{sec:exp}

In this section, we first demonstrate and evaluate the capability of our Changen2 on change data synthesis (Sec.~\ref{sec:cds}).
Based on synthetic change data from our Changen2 model, we carefully evaluate the effectiveness of synthetic data pre-training (Sec.~\ref{sec:sdp}) in three aspects: zero-shot object change detection (Sec.~\ref{sec:sdp:zs}), adaptation for object change detection (Sec.~\ref{sec:sdp:ocd}), and adaptation for semantic change detection (Sec.~\ref{sec:sdp:scd}).
Finally, we compare our Changen2-based synthetic data pre-training with state-of-the-art remote sensing foundation models (Sec.~\ref{sec:rsfm}).

\begin{table}[htb]
\caption{Benchmark comparison of image quality ($256^2$ px) across different deep generative models in the context of change data generation.
\label{tab:gen_model}}
\centering
\small
\tablestyle{2pt}{1.4}
\resizebox{0.98\linewidth}{!}{
\begin{tabular}{l|l|ccc}
Method              &  Modeling   & FID$\downarrow$ & IS$\uparrow$ & Ref. \\
\shline
GPCM \cite{changen}        &              & - & - & ICCV'23 \\
+ SPADE \cite{spade}       & \multirow{4}{*}{$P(\I_t|\Sx{t})$}   & 204.01    &  3.41  & CVPR'19 \\
+ OASIS \cite{oasis}       &                                     &  45.13    &  4.95  & ICLR'21 \\
+ ControlNet (SD 1.5) \cite{controlnet} &   &  101.45  &  6.70    & ICCV'23 \\  
+ ControlNet (SD 2.1) \cite{controlnet} &   &  92.98   &  6.26    & ICCV'23 \\
\hline
+ DiT-B/2 \cite{dit} (our modified) &\multirow{3}{*}{$P(\I_{t+1}|\Sx{t+1}, \I_t)$}&  32.33   &  4.66    & ICCV'23 \\
Changen \cite{changen}  &  &  34.74  & 5.41 & ICCV'23 \\   
Changen2 (ours)&   & 32.44  &  4.64    & - \\
\end{tabular}}
\vspace{-2mm}
\end{table}

\subsection{Change Data Synthesis}\label{sec:cds}

\mpara{0mm}{Setup.}
Following the setting of Changen \cite{changen}, for object change data generation, we use \textit{xView2 pre-disaster} \cite{gupta2019xbd}, which is a globally distributed satellite image dataset with pre-disaster building footprint annotations.
We use \texttt{train} and \texttt{tier3} splits of this dataset to train each generative model.
The \texttt{hold} split, which is cropped into 256$\times$256 non-overlapped patches, is used to evaluate image quality with FID \cite{FID} and IS \cite{IS}.
For semantic change data generation, we use \textit{OpenEarthMap} \cite{oem}, which is also a globally distributed satellite image dataset with land cover annotations.

\begin{figure*}[htb]
\centering
\includegraphics[width=\linewidth]{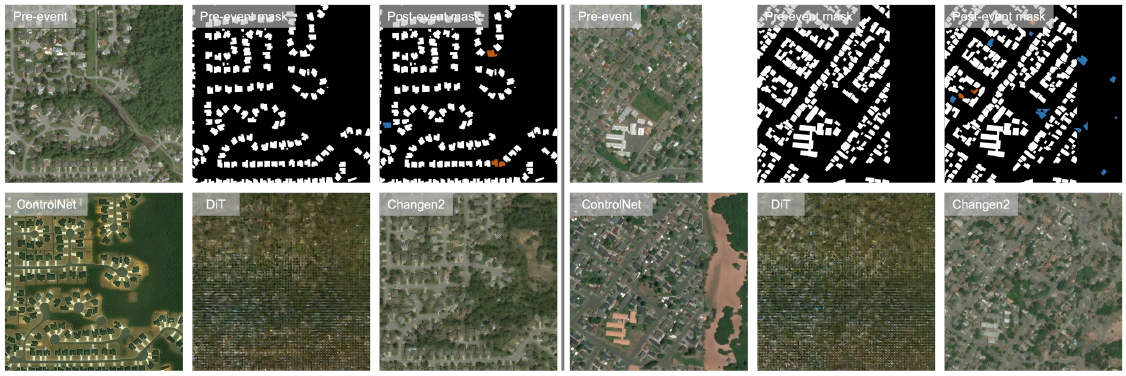}
\caption{\textbf{Spatial Resolution Scalability (256$^2$ px to 1,024$^2$ px)}.
``Pre-event'' presents a real pre-event image.
``Post-event mask'' presents a simulated post-event semantic mask, where its legends are \textcolor{create}{\textbf{Create}} and \textcolor{remove}{\textbf{Remove}}.
The bottom three in a subfigure are synthesized post-event images \textit{without pre-event image guidance}.
\textbf{(Left)} Changen2 generates a more realistic image than SD + ControlNet.
\textbf{(Right)} Changen2 outpaints a more consistent content than SD + ControlNet.
}
\label{fig:256_to_1024}
\vspace{-4mm}
\end{figure*}

\mpara{1mm}{Implementation details.}
Each generative model is trained at 256$\times$256 px if not specified
For GAN-based approaches, 
SPADE \cite{spade}, OASIS \cite{oasis}, Changen \cite{changen} are trained with Adam ($\beta_1=0$, $\beta_2=0.999$), a batch size of 32, and 100k iterations
The learning rate is 0.0001 for the generator and 0.0004 for the discriminator.
For DPM-based approaches,
ControlNets \cite{controlnet} are based on pre-trained Stable Diffusion (SD) \cite{rombach2022high}, where SD part is frozen.
DiT \cite{dit} and our Changen2 are trained from scratch.
DiT is modified with our proposed dense embedding network to support dense conditions.
These diffusion models are trained with AdamW \cite{loshchilov2017decoupled} ($\beta_1=0.9$, $\beta_2=0.999$), weight decay of 0, learning rate of 0.0001, a batch size of 32, and 500k iterations.
All diffusion models are trained in VAE's latent space.
We adopt an off-the-shelf pre-trained variational autoencoder (VAE) model \cite{kingma2013auto} from SD \cite{rombach2022high}.
The VAE is frozen, responsible only for encoding images into latent space and decoding latents back into image space.

\begin{figure}[tb]
\centering
\includegraphics[width=\linewidth]{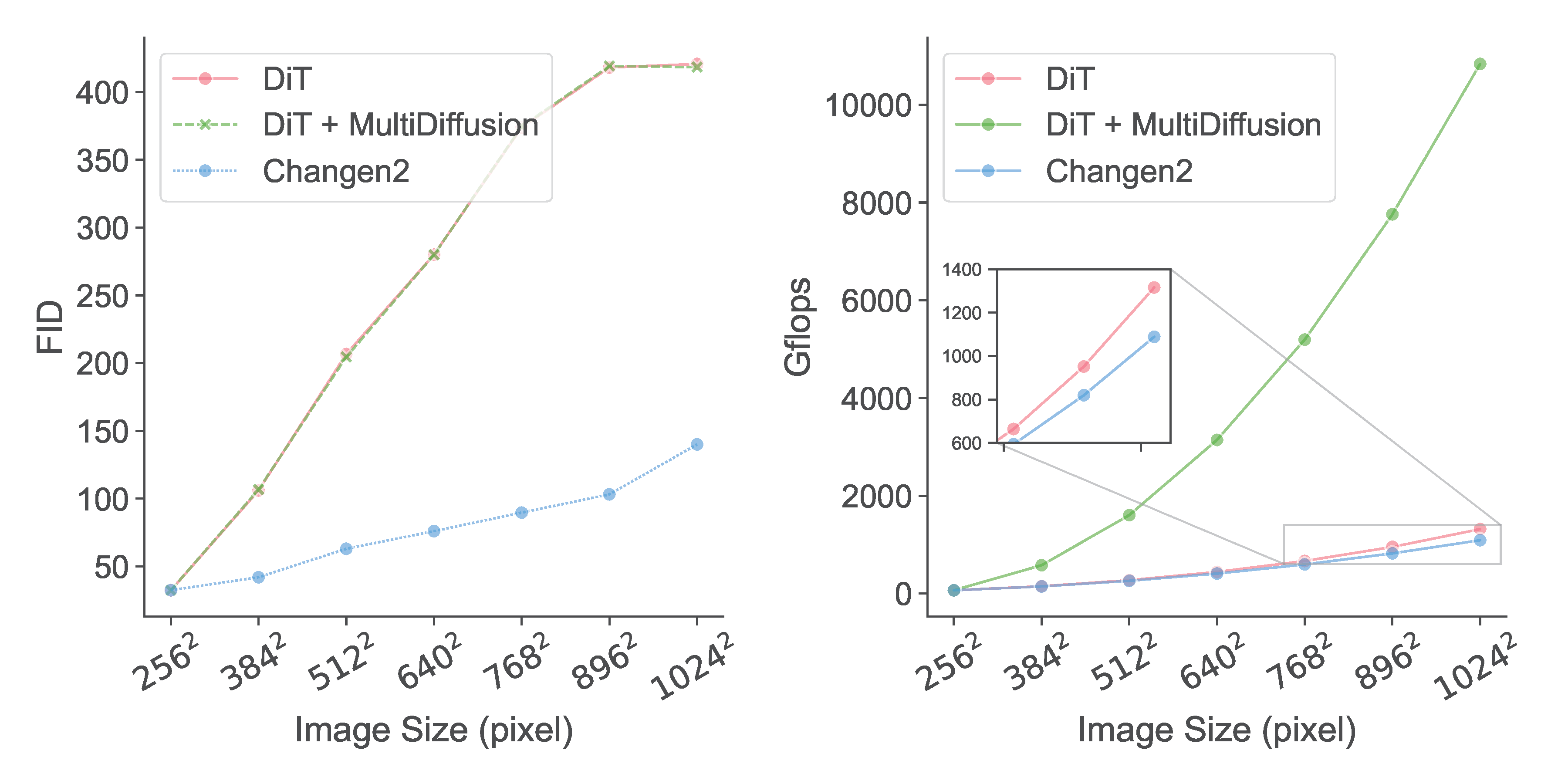}
\caption{\textbf{Quantitative Results of Scalability on Spatial Resolution}.
Changen2 has lower computational complexities and can generate higher-quality images as spatial resolution increases.
}
\label{fig:resolution}
\end{figure}

\mpara{1mm}{Result: Image Quality.}
Table~\ref{tab:gen_model} presents the results of image quality evaluations for various deep generative models within the GPCM framework for change data generation.
For GAN-based models, we compare two representative models: SPADE \cite{spade} and OASIS \cite{oasis}, where OASIS is the baseline of Changen \cite{changen}.
For DM-based models,  we compare state-of-the-art ControlNet \cite{controlnet} with pre-trained SD \cite{rombach2022high} and DiT \cite{dit}.
We have following observations:
(i) Changen2 improves over Changen with higher FID, thanks to the advantage of diffusion models and transformer architecture.
(ii) Changen2 and our modified DiT perform similarly, outperforming ControlNet with SD 1.5 and SD 2.1 and other models, in terms of FID.
(iii) ControlNet with SD does not work well in terms of quantitative FID evaluation.
We argue that this is because the SD part is mainly pre-trained on natural/web images at scale.
The benefit of pre-trained SD is reflected at higher IS.
However, fine-tuning ControlNet for satellite images remains challenging due to the persistent domain gap, especially statistic bias.

\begin{figure*}[htb]
\centering
\includegraphics[width=\linewidth]{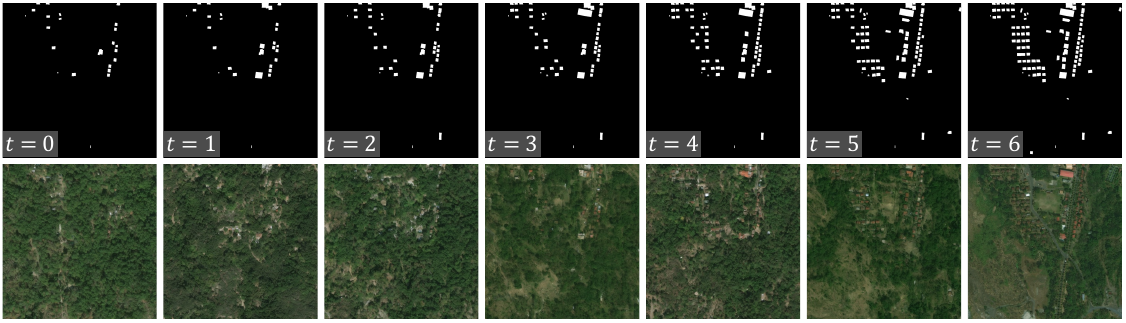}
\caption{\textbf{Spatiotemporal Resolution Scalability: a case of ``Rural Revitalization'' simulation}.
All synthesized post-event images have 1,024$\times$1,024 px and are generated by Changen2 trained at 256$\times$256 px.
}
\label{fig:bulding_ts}
\vspace{-4mm}
\end{figure*}

\begin{figure*}[htb]
\centering
\includegraphics[width=\linewidth]{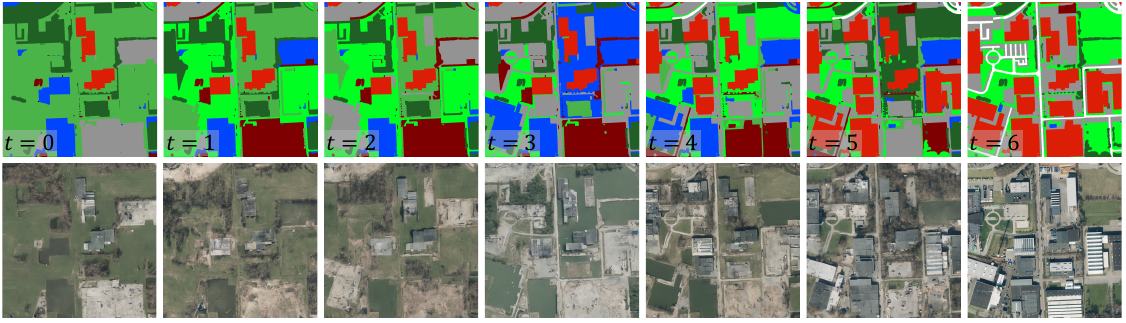}
\caption{\textbf{Spatiotemporal Resolution Scalability: a case of ``Urbanization'' simulation}.
All synthesized post-event images have 1,024$\times$1,024 px and are generated by Changen2 trained at 256$\times$256 px.
}
\label{fig:landcov_ts}
\vspace{-4mm}
\end{figure*}

\begin{figure*}[!htb]
\centering
\includegraphics[width=\linewidth]{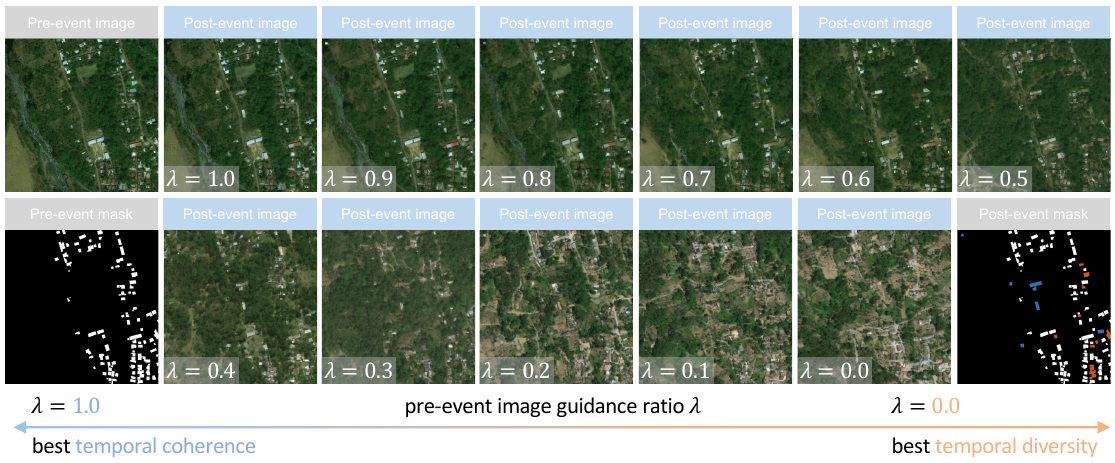}
\caption{\textbf{Trade-off between temporal coherence and temporal diversity}.
Changen2 is capable of generating more temporally coherent (larger $\lambda$) or more temporally diverse (smaller $\lambda$) post-event images by adjusting the pre-event image guidance ratio $\lambda \in [0, 1]$.
All synthesized post-event images have 1,024$\times$1,024 px and are generated by Changen2 trained at 256$\times$256 px.
}
\label{fig:coh_div}
\vspace{-4mm}
\end{figure*}

\mpara{1mm}{Scalability in Spatial Resolution.}
Fig.~\ref{fig:256_to_1024} demonstrates the scalability in the spatial resolution for ControlNet, DiT, and our Changen2, qualitatively.
These three models are trained at 256$^2$ px and generate post-event images at 1,024$^2$ px.
We observe that DiT fails to generate normal images, indicating a lack of scalability in spatial resolution.
ControlNet, due to its convolutional architecture, naturally excels at generating larger images.
However, its generated images lack diversity, i.e., in Fig.~\ref{fig:256_to_1024}-left, all buildings have a similar appearance, and the texture of vegetation is over-smooth.
In the outpainting case shown in Fig.~\ref{fig:256_to_1024}-right, ControlNet generates spatially inconsistent image content due to the locality of major convolution layers. 
Thanks to resolution-scalable transformer architecture, Changen2 can generate more realistic and spatially consistent post-event images.

\mpara{1mm}{Analysis and Ablation: Spatial Resolution Scalability.}
To investigate the underlying reasons for Changen2's good scalability in spatial resolution, we conduct a step-by-step analysis, from DiT to Changen2.
From the architecture perspective, only absolute position embedding limits DiT's input size in principle.
Thus, we interpolate its position embedding or re-generate its position embedding by new size, as shown in Fig.~\ref{fig:256_to_512}, these two solutions do not work. 
Using MultiDiffusion \cite{multidiffusion}, a powerful training-free adaptation algorithm capable of generating higher-resolution images from low-resolution diffusion models, DiT generates a normal image. 
However, the image content is misaligned with the semantic mask, exhibiting issues such as inconsistent spatial layout and the absence of buildings (incorrect semantics).
Re-training a DiT model at 512$^2$ px works well for 512$^2$ px image generation, however, its training time is much longer than a model trained at 256$^2$ px.
Our Changen2 model, trained at 256$^2$ px, performs well for generating images at 512$^2$ px and even at 1,024$^2$ px.
There are only two small but important improvements over DiT: (a) removing absolute position embedding and adding a 3$\times$3 depthwise convolution layer in each block; (b) replacing global self-attention with local window-attention.
We ablate these two improvements step-by-step in Fig.~\ref{fig:256_to_512}.
The results suggest that the improvement (a) effectively addresses the issue of spatial resolution scalability in DiT, confirming that the root obstacle is the absolute position embedding.

We also quantitatively evaluate the performance of DiT, DiT with MultiDiffusion, and our Changen2 for scalability in spatial resolution, as shown in Fig.~\ref{fig:resolution}
Changen2 significantly improves over both DiT and DiT with MultiDiffusion in terms of image quality while exhibiting the lowest computational complexity, with a reduction of up to 17\%.

\mpara{1mm}{Scalability in Spatiotemporal Resolution.}
Changen2 inherits from Changen's temporal resolution scalability, as it remains within the our GPCM framework.
Therefore, Changen2 has spatiotemporal resolution scalability.
As shown in Fig.~\ref{fig:bulding_ts} and Fig.~\ref{fig:landcov_ts}, Changen2 demonstrates good scalability in both spatial and temporal resolution for generating building change data and land cover change data, respectively.
These two cases, "Rural Revitalization" and "Urbanization," further highlight Changen2's highly customizable capabilities for simulating change events.

\mpara{1mm}{Trade-off between Temporal Coherence and Diversity.}
Changen2 can flexibly control temporal coherence and diversity to meet various application scenarios by adjusting the pre-event image guidance ratio $\lambda$ during inference.
As shown in Fig.~\ref{fig:coh_div}, we show a case that gradually reduces $\lambda$ for the post-event image generation.
As expected, a larger $\lambda$ results in a more consistent spatial layout and object appearance, providing better temporal coherence between the pre-event and post-event images.
This is useful in high-fidelity simulation scenarios, e.g., assisting urban planning in a virtual environment.
A smaller $\lambda$ results in a more diverse spatial layout and object appearance, providing harder and more informative positive and negative examples for synthetic change data pre-training.
Temporal diversity is important for synthetic data pre-training.

\subsection{Synthetic Data Pre-training}\label{sec:sdp}

\noindent\textbf{Implmentation details.}
To evaluate the effectiveness of our synthetic data pre-training, we generate a building change detection dataset \texttt{Changen2-S1-15k} from \textit{xView2 pre-disaster}, a semantic change detection \texttt{Changen2-S9-27k} from \textit{OpenEarthMap}, and a class-agnostic change detection dataset \texttt{Changen2-S0-1.2M} from unlabeled fMoW satellite images.
The dataset name follows a template of \texttt{Changen2-S<{\rm number of known object classes}>-<{\rm number of image pairs}>}, e.g., \texttt{Changen2-S1-15k} means that this dataset includes one known object class (i.e., building) and 15k bitemporal image pairs.
Each image has 512$\times$ 512 px.
We pre-train a ChangeStar \cite{changestar} model, a simple yet effective multi-task change detection architecture, on each dataset individually.
The ChangeStar newly trained in this work adopts an optimized configuration of FarSeg++ \cite{zheng2023farseg++} with 96 or 256 channels, ChangeMixin with $N = 1$, $d_c = 96$ or $256$, which denotes ChangeStar (1$\times$96 or 1$\times$256).
The pre-training follows standard bitemporal supervised learning.
If not specified, AdamW is used as the optimizer with a weight decay of 0.01.
The total batch size is 16.
The initial learning rate is 0.0001 with a ``poly''($\gamma=0.9$) decay scheduler. 
We train the model with 40k, 80k, and 800k iterations on \texttt{Changen2-S1-15k}, \texttt{Changen2-S9-27k}, and \texttt{Changen2-S0-1.2M}, respectively, resulting in three pre-trained models.
Only D4 dihedral group transformations are used for training data augmentation.

\vspace{-1mm}
\subsubsection{Zero-shot Change Detection}\label{sec:sdp:zs}

\mpara{0mm}{Setup}.
The zero pretext task gap nature of synthetic data pre-training distinguishes it from other pre-training technologies.
This characteristic enables the change detector pre-trained on synthetic change data to perform zero-shot change detection.
As defined in \cite{anychange}, zero-shot capability indicates that the entire pipeline (comprising a change detector and a change data generator) can generalize to unseen tasks and data distributions. 
Each pipeline has only seen real single-image segmentation tasks, such as building segmentation in xView2 pre-disaster. 
Its change data generator converts these single-image segmentation tasks into multi-temporal change detection tasks for training its change detector.
Following the protocol in \cite{changen}, we evaluate pipelines on the entire LEVIR-CD \cite{levircd} and WHU-CD \cite{whucd} datasets.
These two datasets do not geographically overlap with xView2, therefore, they can be considered as unseen data.
To ensure a fair evaluation of change data generators, we utilize the same change detection architecture—ChangeStar (1$\times$96) with ResNet-18 as the backbone.
We employ a rigorous and challenging generalization evaluation protocol in which \textbf{a single model}, pre-trained on synthetic change data, is evaluated across multiple datasets.

\begin{table}[t]
\caption{\textbf{Zero-shot Object Change Detection}.
Comparison with other data generators on LEVIR-CD$^\texttt{all}$ and WHU-CD$^\texttt{all}$.
``\Create'' and ``\Remove'' denote creating objects and removing objects, respectively.
All change detection architectures are ChangeStar (1$\times$96) based on ResNet-18 for a consistent and fair comparison.
\label{tab:zero}}
\vspace{-1em}
\centering
\small
\resizebox{1.\linewidth}{!}{
\tablestyle{1.5pt}{1.4}
\begin{tabular}{l|c|cccc|cccc}
               & Supported        & \multicolumn{4}{c|}{LEVIR-CD$^{\texttt{all}}$} & \multicolumn{4}{c}{WHU-CD$^{\texttt{all}}$}                                             \\
Data generator   & event type       & IoU & F$_1$ & Prec. & Rec. & IoU & F$_1$ & Prec. & Rec. \\
\shline
Copy-Paste     & \Create       & 0.7  &  1.4 & 1.9   & 1.2  & 1.9 & 3.8   & 2.4   & 9.0  \\
Inpainting     & \Remove       & 10.8 & 19.5 & 11.9  & 54.0 & 12.2& 21.8  & 13.3  & 59.3 \\
OASIS+GPCM & \Create\&\Remove  & 39.7  & 56.8 & 45.5  & 75.7 &  12.7 & 22.6 & 13.8  & 61.3 \\
ControlNet+GPCM & \Create\&\Remove & 44.0 & 61.1 & 48.5 & 82.5 & 16.4 & 28.2 & 16.9 & 84.9 \\ 
DiT-B/2+GPCM & \Create\&\Remove & 68.2 & 81.1 & 80.0 & 82.2 & 58.1 & 73.5 & 68.3 & 79.6  \\
\hline
Changen & \Create\&\Remove & 45.8 & 62.8 & 49.3 & \best{86.4} & 15.3 & 26.6 & 15.7  & \best{87.1} \\
Changen2 (ours) & \Create\&\Remove & \best{72.0} & \best{83.7} & \best{87.1} & 80.6 & \best{61.7} & \best{76.3} & \best{75.5} & 77.1 \\
$\Delta$ gains              &    &   & \up{20.9} &   &   &   &\up{49.7}   \\
\hline
\textit{architecture improvement}          &             &            &   \\
+ {\scriptsize ChangeStar w/ MiT-B1} \text & \Create\&\Remove & 75.9 & 86.3 & 82.0 & 91.1 & 70.8 & 82.9 & 79.2 & 86.9 \\
+ {\scriptsize ChangeStar w/ SwinV2-B} \text & \Create\&\Remove & 74.5 & 85.4 & 79.9 & 91.7 & 65.5 & 79.1 & 72.1  & 87.8 \\
+ {\scriptsize ChangeStar w/ SAM/ViT-B} \text & \Create\&\Remove & 80.0 & 88.9  & 87.8 & 89.9  & 71.0 & 83.1 & 78.8 & 87.9 \\
+ {\scriptsize ChangeStar w/ SAM/ViT-L} \text & \Create\&\Remove & 81.5 & 89.8  & 87.5 & 92.3  & 72.4 & 84.0 & 79.3 & 89.2 \\
\end{tabular}}
\vspace{-2mm}
\end{table}

\mpara{0.5mm}{Result: comparison with other data generators}.
Table~\ref{tab:zero} suggests that Changen2 outperforms other counterparts including competitive ControlNet and DiT.
Compared with the previous version, i.e., Changen, Changen2 improves over it by a significant margin, where 20.9\% and 49.7\% F$_1$ gains are obtained on LEVIR-CD and WHU-CD datasets.
This suggests that Changen2 generates more accurate synthetic change data. Specifically, the bitemporal images correspond much better with their change masks, as Changen2 fundamentally avoids feature leakage problems.

\begin{table}[t]
\caption{\textbf{Adaptation for Object Change Detection}. Comparison with the state-of-the-art change detectors on \textbf{LEVIR-CD} \texttt{test}.
``R-18'': ResNet-18.
$\dagger$ indicates that the backbone is pre-trained on ImageNet and then ADE20K \cite{ade20k}.
$*$ indicates their modified backbone. 
The amount of floating point operations (Flops) was computed with a float32 tensor of shape [2,256,256,3].
ChangeStar (1$\times$256) with ViT-L has 318.5M parameters and the Flops of 288.3G. 
\label{tab:bench_levir}}
\vspace{-1em}
\centering
\small
\tablestyle{1.pt}{1.4}
\resizebox{\linewidth}{!}{
\begin{tabular}{l|c|c|l|cc}
Architecture & Pre-trained from &Backbone & F$_1\uparrow$ & \#Params. & \#Flops \\
\shline
\ft\textit{fully-supervised} &&&&& \\
ChangeStar  \cite{changestar} & ImageNet-1K   & R-18 & 90.2  & 19.3M&  22.3G   \\
ChangeStar  \cite{changestar}& ImageNet-1K   & R-50 & 90.8  & 33.9M &  29.0G    \\
ChangeStar  \cite{changestar}& ImageNet-1K   & RX-101 & 91.2 & 52.5M & 39.2G    \\
ChangeFormer \cite{changeformer} & IN-1K,ADE20K$^\dagger$ & MiT-B2$^*$ & 90.4 & 41.0M & 203.1G \\
BiT \cite{chen2021remote,wang2022empirical} & ImageNet-1K & {\scriptsize ViTAEv2-S} &  91.2       &  19.6M   & 15.7G    \\
BAN-BiT \cite{li2023new}& CLIP \cite{clip} & ViT-L/14 & 91.9 & 231.7M & 298.2G     \\
\hline
ChangeStar (1$\times$96)  & ImageNet-1K & R-18  & 90.5 & 16.4M      &  16.3G  \\
+ self-supervised         &  SeCo-1M \cite{seco}  & R-18 & 89.9 & 16.4M &  16.3G  \\
+ self-supervised         &  MoCov2 \cite{mocov2} & R-18& 90.4 & 16.4M &  16.3G  \\
+ seg. supervised         & xView2 pre-disaster   & R-18 & 90.6 & 16.4M      &  16.3G  \\
+ OASIS+GPCM \cite{changen} & \texttt{OASIS-90k}  & R-18 & 90.6 & 16.4M &  16.3G  \\
+ Changen \cite{changen} & \texttt{Changen-90k}   & R-18 & 91.1   & 16.4M &  16.3G  \\
+ DiT-B/2+GPCM \cite{changen} & \texttt{DiT-B/2-15k} & R-18 & 90.5 & 16.4M &  16.3G  \\
\gr + Changen2 (ours) & \texttt{Changen2-S1-15k} & R-18  & 91.3 & 16.4M &  16.3G  \\
\hline
ChangeStar (1$\times$96)  &  ImageNet-1K  & MiT-B1 & 90.0 & 18.4M &  16.0G        \\
+ Changen \cite{changen}  &  \texttt{Changen-90k}  &  MiT-B1  & 91.5 &  18.4M    &  16.0G \\
\gr + Changen2 (ours)     &  \texttt{Changen2-S1-15k}&  MiT-B1 & 91.9 &  18.4M  & 16.0G \\ 
\hline
ChangeStar (1$\times$256) & Satlas-HiRes \cite{satlas} & SwinV2-B &  91.7 & 96.9M  &  63.0G \\
\gr + Changen2 (ours)     &  \texttt{Changen2-S1-15k}& SwinV2-B & 92.0 & 96.9M  &  63.0G \\
\hline
ChangeStar (1$\times$256) & SA-1B \cite{sam} &ViT-B& 92.0 &  99.4M & 93.3G \\
\gr + Changen2 (ours)     & \texttt{Changen2-S1-15k}&ViT-B& 92.5 & 99.4M & 93.3G \\
\shline
\zs\textit{\textcolor{olive}{zero-shot}} &&&&fully sup.& F$_1$ gap \\
ChangeStar (1$\times$96)&\texttt{Changen2-S1-15k}&R-18&83.7 &91.3& -7.6 \\
ChangeStar (1$\times$96)&\texttt{Changen2-S1-15k}&MiT-B1&86.4&91.9&-5.5 \\
ChangeStar (1$\times$256)&\texttt{Changen2-S1-15k}&SwinV2-B&85.4&92.0&-6.6 \\
ChangeStar (1$\times$256)&\texttt{Changen2-S1-15k}&ViT-B&88.7&92.5&-3.8 \\
ChangeStar (1$\times$256)&\texttt{Changen2-S1-15k}&ViT-L&89.6& - &-2.9\\
\end{tabular}}
\vspace{-4mm}
\end{table}

\mpara{1mm}{Result: zero-shot object change detection on LEVIR-CD}.
Table~\ref{tab:bench_levir} presents that ChangeStar models pre-trained on our \texttt{Changen2-S1-15k} datasets exhibit outstanding zero-shot performance, comparable to fully supervised counterpart on LEVIR-CD benchmark for the first time.
We observe that a larger model yields better zero-shot performance and a smaller F$_1$ gap between the zero-shot model and its fully supervised counterpart.
For example, the zero-shot R-18 variant achieves 83.7\% F$_1$ with a -7.6\% gap, while the zero-shot ViT-B variant achieves 88.7\% F$_1$ with only a -3.8\% gap.
Using the ViT-B variant as a reference, the zero-shot ViT-L variant achieves 89.6\% F$_1$, further reducing the gap to -2.9\%.

\begin{table}[t]
\caption{\textbf{Adaptation for Object Change Detection}. Comparison with the state-of-the-art change detectors on the \textbf{S2Looking} \texttt{test} set.
``R-18'': ResNet-18.
The amount of floating point operations (Flops) was computed with a float32 tensor of shape [2,512,512,3] as input.
ChangeStar (1$\times$256) with ViT-L has 318.5M parameters and the Flops of 914.0G. 
\label{tab:bench_s2l}}
\centering
\small
\tablestyle{2.5pt}{1.4}
\resizebox{\linewidth}{!}{
\begin{tabular}{l|cc|ccc|cc}
 Architecture                           & Pre-trained from  & Backbone        & F$_1\uparrow$         & Prec. & Rec. & \#Params. & \#Flops \\
 \shline
 \ft\textit{fully-supervised} &&&&&&& \\
FC-Siam-Diff \cite{daudt2018fully}  & -  & -   & 13.1          & 83.2  & 15.7 & 1.3M       & 18.7G   \\
STANet \cite{levircd}             &ImageNet-1K & R-18            & 45.9          & 38.7  & 56.4 & 16.9M      & 156.7G  \\
RDP-Net \cite{rdpnet} & -  & - &60.5  &  65.9  &  55.9   & 1.7M & 108.5G  \\
BAN-BiT \cite{li2023new} & CLIP \cite{clip} & ViT-L/14 & 65.4 & 75.1  & 58.0 & 231.7 M & 330.2G \\
A2Net \cite{a2net}    & ImageNet & R-18  & 66.3 &  69.3  &   63.6  & 17.4M & 55.2G \\
\hline
ChangeStar (1$\times$96) &ImageNet-1K & R-18 & 66.3 & 70.9  & 62.2 & 16.4M & 65.3G   \\
+ Changen \cite{changen} &\texttt{Changen-90k} & R-18 & 67.1& 70.1& 64.3 & 16.4M & 65.3G \\
\gr + Changen2 (ours) &\texttt{Changen2-S1-15k}& R-18 &67.3&72.0&63.1&16.4M&65.3G \\
\hline
ChangeStar (1$\times$96)&ImageNet-1K&MiT-B1& 64.3 & 69.3  & 59.9  & 18.4M & 67.3G\\
+ Changen \cite{changen}&\texttt{Changen-90k} & MiT-B1&67.9&70.3&65.7&18.4M&67.3G\\
\gr + Changen2 (ours) &\texttt{Changen2-S1-15k} & MiT-B1 &68.4&71.9&65.2&18.4M&67.3G\\
\hline  
ChangeStar (1$\times$256)&Satlas-HiRes \cite{satlas}&SwinV2-B&68.3&72.6&64.5&96.9M&251.8G\\
\gr + Changen2 (ours) &\texttt{Changen2-S1-15k}&SwinV2-B&68.9&71.3&66.6& 96.9M&251.8G\\
\hline
ChangeStar (1$\times$256) & SA-1B \cite{sam} &ViT-B&  68.1  & 71.5  & 64.9 & 99.4M & 321.4G \\
\gr + Changen2 (ours) &\texttt{Changen2-S1-15k} &ViT-B&68.5&74.3&63.6& 99.4M & 321.4G \\
\shline
\zs\textit{\textcolor{olive}{zero-shot}} &&&&&&fully sup.& F$_1$ gap \\
ChangeStar (1$\times$96)&\texttt{Changen2-S1-15k}&R-18&35.8&36.1&35.4&67.3&-31.5\\
ChangeStar (1$\times$96)&\texttt{Changen2-S1-15k}&MiT-B1&41.7&41.0&42.3&68.4&-26.7 \\
ChangeStar (1$\times$256)&\texttt{Changen2-S1-15k}&SwinV2-B&45.6&46.2&45.0&68.9&-23.3 \\
ChangeStar (1$\times$256)&\texttt{Changen2-S1-15k}&ViT-B&44.1&56.1&36.3&68.5&-24.4 \\
ChangeStar (1$\times$256)&\texttt{Changen2-S1-15k}&ViT-L&52.7&55.5&50.1& - & -16.2 \\
\end{tabular}}
\vspace{-2mm}
\end{table}

\begin{figure}[t]
\centering
\includegraphics[width=\linewidth]{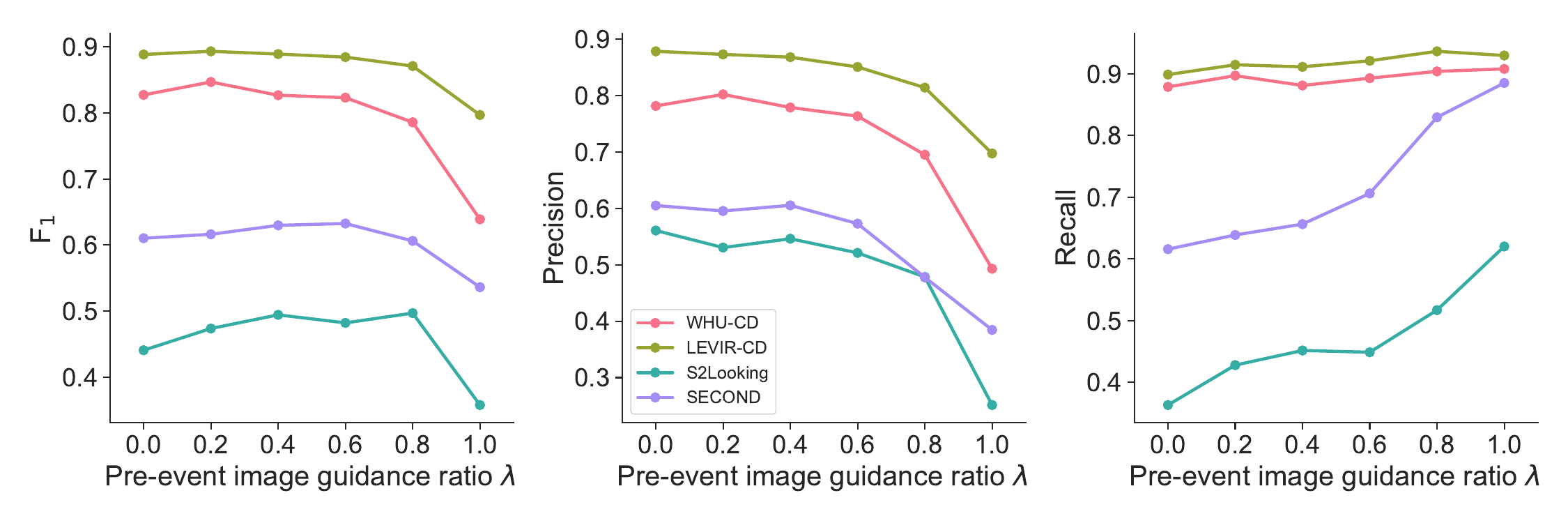}
\caption{\textbf{Ablation: how pre-event image guidance ratio $\lambda$ impact zero-shot performance}.
Temporal diversity (smaller $\lambda$) is generally important to zero-shot performance.
}
\label{fig:lambda}
\vspace{-4mm}
\end{figure}

\mpara{1mm}{Result: zero-shot object change detection on S2Looking}.
Evaluating the same single models on S2Looking, Table~\ref{tab:bench_s2l} (zero-shot part) also presents similar observations, i.e., larger models benefit more from our synthetic change data, exhibiting higher zero-shot performance and smaller F$_1$ gap.
Even in more challenging off-nadir scenarios in the S2Looking benchmark, the F$_1$ gap can be reduced to -16.2\% using \texttt{Changen2-S1-15k} for the first time.

\mpara{1mm}{Result: zero-shot change detection on SECOND}.
A similar observation can be found in the SECOND benchmark (Table~\ref{tab:bench_scd}, zero-shot), which includes up to 30 change types, making it more complex than the LEVIR-CD and S2Looking benchmarks in terms of the variety of change types.
ChangeStar (1$\times$256) pre-trained on \texttt{Changen2-S9-27k} with ViT-B, achieves 61.0\% F$_1$, reducing the gap with fully supervised counterpart to -12.6\% F$_1$.
Compared to the previous state-of-the-art AnyChange \cite{anychange} at the system level, our model improves pixel-level zero-shot performance by 16.4\% F$_1$ while reducing computational complexity to 80.7\%.

\mpara{1mm}{Ablation: Pre-event image guidance ratio $\lambda$}.
Fig.~\ref{fig:lambda} presents zero-shot performances using varying $\lambda$.
We can observe that smaller $\lambda$ always results in better zero-shot F$_1$ scores across four datasets, which means that temporal diversity of synthetic change data is the key to zero-shot performance.
Larger $\lambda$ results in higher recall and smaller $\lambda$ results in higher precision.
These results clearly reveal, for the first time, the relationship between zero-shot performance, temporal diversity, and temporal coherence.
We can use these empirical rules to adjust our synthetic change data for different application scenarios (e.g., higher recall for disaster response).

\mpara{1mm}{Observation: Scaling up change detector benefits more from our synthetic change data}.
With the same synthetic change data \texttt{Changen2-S1-15k} or \texttt{Changen2-S9-27k},
the larger model can exhibit better zero-shot performance and a smaller gap with its fully supervised counterpart.
These observations (Table~\ref{tab:bench_levir},~\ref{tab:bench_s2l},~\ref{tab:bench_scd}) demonstrate the scalability of Changen2 pre-training.

\subsubsection{Adaptation for Object Change Detection}\label{sec:sdp:ocd}
\mpara{0mm}{Setup}.
As is common with conventional pre-trained models, it is standard practice to fine-tune them on downstream task data to fully explore their potential \cite{mae,satmae}.
We also adapt our Changen2 pre-trained models to object change detection by fully supervised fine-tuning on LEVIR-CD and S2Looking datasets.
By default, we use AdamW as the optimizer with a weight decay of 0.01, an initial learning rate of 6e-5 with ``poly'' ($\gamma=0.9$) decay.
Each model is trained with 20k steps with a total batch size of 16.
Training data augmentation adopts D4 dihedral group transformations, scale jitter, and random cropping into 512$\times$512.
For evaluation, we directly apply the trained model for inference on bitemporal images of 1,024$\times$1,024 px, without any tricks.

\mpara{1mm}{Result on LEVIR-CD}.
Table~\ref{tab:bench_levir} presents a comprehensive benchmark for regular building change detection.
We argue data matters.
Using the same architecture, ChangeStar (1$\times$96), 
Changen2 pre-training outperforms previous state-of-the-art supervised pre-training (SA-1B \cite{sam} and Satlas pre-training \cite{satlas}).
Compared to Changen and DiT+GPCM pre-training, Changen2 pre-training offers better performance gain to the same architecture.
\texttt{Changen2-S1-15K} includes xView2 pre-disaster data, thus, we pre-trained a model with this data to ablate its performance gain.
We find that Changen2 pre-training is still superior to xView2 pre-disaster data pre-training, which confirms Changen2 brings information gain via simulating change events.
More importantly, for the first time, the zero-shot performance brought by Changen2 pre-training is comparable to fully supervised counterparts.
This confirms the transferability of Changen2 pre-training and the feasibility of our roadmap for zero-shot change detection, i.e., synthetic data pre-training.

\mpara{1mm}{Result on S2Looking}.
Table~\ref{tab:bench_s2l}  presents a comprehensive benchmark for off-nadir building change detection.
Changen2 pre-training outperforms Satlas \cite{satlas} and SA-1B pre-training by 0.6\% and 0.4\% in F$_1$ score, respectively. 
Besides, it consistently improves over Changen pre-training.
For a system-level comparison, our ChangeStar (1$\times$256) with ViT-B achieves 68.5\% F$_1$, with fewer parameters and Flops, outperforming BAN-BiT \cite{li2023new} with CLIP \cite{clip} pre-trained ViT-L by 2.2\%.
Additionally, we observe that our zero-shot ChangeStar (1$\times$256) with ViT-L outperforms two early classical fully supervised models, namely FC-Siam-Diff and STANet.
These results further support our claim that data matters and validate the feasibility of our synthetic data pre-training roadmap for zero-shot change detection.

\begin{table*}[t]
  \caption{\textbf{Adaptation for Semantic Change Detection}. 
  Comparison with state-of-the-art semantic change detection models on official SECOND \texttt{test}.
  The amount of floating point operations (Flops) was computed with a float32 tensor of shape [2,512,512,3] as input.
  \label{tab:bench_scd}}
  \centering
  \small
  \tablestyle{4pt}{1.4}
  \begin{tabular}{l|cc|ccc|cc|cc}
 \multirow{2}{*}{Architecture}&\multirow{2}{*}{Pre-training data}&\multirow{2}{*}{Backbone}&\multicolumn{3}{c|}{BCD} & \multicolumn{2}{c|}{SCD} &\multirow{2}{*}{\#Params.}    & \multirow{2}{*}{\#Flops}  \\
&    &   & F$_1\uparrow$ & Prec. & Rec. & SeK$_{37}\uparrow$ & mIoU$_{30}\uparrow$&  &  \\
\shline
\ft\textit{fully-supervised} &&&&&&&&& \\
Siamese-UNet&ImageNet-1K&EfficientNet-B0&71.6&72.1&71.0&18.6&26.0& 9.6M & 35.7G\\
Bi-SRNet\cite{ding2022bi}& ImageNet-1K& R-34 &72.5&73.2&71.7&19.8&26.1&23.4M&199.5G\\
ChangeMask\cite{zheng2022changemask}&ImageNet-1K&EfficientNet-B0&72.0&75.5&68.8&19.3&27.4&10.6M& 38.1G\\
\hline
ChangeStar (1$\times$96)&ImageNet-1K&EfficientNet-B0&71.3&74.0&68.8&18.9&25.8&9.4M& 58.1G \\
ChangeStar (1$\times$96)&OpenEarthMap\cite{oem}&EfficientNet-B0&71.5&71.6&71.4&19.0&{\bf 27.5}&9.4M& 58.1G\\
\gr + Changen2 (ours) &\texttt{Changen2-S9-27k}&EfficientNet-B0&{\bf 72.1}&74.1&70.1&{\bf 20.2}&26.2& 9.4M& 58.1G\\
\hline
ChangeStar (1$\times$256)&ImageNet-1K&SwinV2-B&71.9&76.5&67.8&20.3&27.1&96.9M& 271.2G \\
ChangeStar (1$\times$256)&OpenEarthMap\cite{oem}&SwinV2-B&72.5&75.6&69.6&20.8&28.7&96.9M& 271.2G \\
ChangeStar (1$\times$256)&Satlas-HiRes \cite{satlas}&SwinV2-B&72.8&77.9&68.3&21.8&28.1&96.9M& 271.2G \\
\gr + Changen2 (ours)&\texttt{Changen2-S9-27k}&SwinV2-B&{\bf 73.0}&72.8&73.3&{\bf 22.6}&{\bf 29.5}&96.9M& 271.2G \\
\hline
ChangeStar (1$\times$256)&OpenEarthMap\cite{oem}&ViT-B&71.6&78.7&65.6&21.2&27.7&99.4M&348.6G \\
ChangeStar (1$\times$256)&SA-1B \cite{sam}&ViT-B&72.8&74.1&71.4&21.5&30.0&99.4M& 348.6G \\
\gr + Changen2 (ours)&\texttt{Changen2-S9-27k}&ViT-B&{\bf 73.6}&76.6&70.8&{\bf 23.0}&{\bf 30.4}&99.4M& 348.6G \\
\shline
\zs\textit{\textcolor{olive}{zero-shot}} &&&&&&fully sup. F$_1$&F$_1$ gap&& \\
DINOv2 + CVA \cite{anychange}&142M images \cite{dinov2}& ViT-G/14& 41.4&26.9& 89.4&-&-&1.1B&3567.9G \\
AnyChange \cite{anychange} &SA-1B \cite{sam}& ViT-B&44.6&30.5&83.2&69.5&-24.9&93.7M&1806.5G \\
\gr ChangeStar (1$\times$96)&\texttt{Changen2-S9-27k}&EfficientNet-B0&57.9&62.2&54.2&72.1&-14.2&9.4M&58.1G\\
\gr ChangeStar (1$\times$256)&\texttt{Changen2-S9-27k}&SwinV2-B&58.2&61.4&55.3&73.0&-14.8&96.9M& 271.2G\\
\gr ChangeStar (1$\times$256)&\texttt{Changen2-S9-27k}&ViT-B&61.0&60.5&61.6&73.6&-12.6&99.4M&348.6G \\
\end{tabular}
\vspace{-3mm}
\end{table*}

\subsubsection{Adaptation for Semantic Change Detection}\label{sec:sdp:scd}
\mpara{0mm}{Setup.}
We further fine-tune the models pre-trained on \texttt{Changen2-S9-27k} on semantic change detection (SCD) task using SECOND dataset \cite{second}.
Compared to object change detection, it involves more complex up to 30 change types.
By default, we use AdamW as the optimizer with a weight decay of 0.01, an initial learning rate of 6e-5 with ``poly'' ($\gamma=0.9$) decay.
Each model is trained with 20k steps with a total batch size of 16.
Training data augmentation adopts D4 dihedral group transformations, scale jitter, and random cropping into 256$\times$256.
For evaluation, we directly apply the trained model for inference on bitemporal images of 512$\times$512 px, without any tricks.

\mpara{1mm}{Metrics.}
The metrics used for semantic change detection is a bit complex than object change detection.
Following standard metrics \cite{second} and common practice \cite{zheng2022changemask, ding2022bi}, we assess the part of class-agnostic change detection (also commonly referred to as binary change detection, BCD) using F$_1$, precision (Prec.), and recall (Rec.).
We assess semantic change detection with an official metric, Separated kappa coefficient (SeK$_{37}$) \cite{second}, where we highlight correct SeK is computed over 37 change types\footnote{Offical implementation can be found here \url{https://captain-whu.github.io/SCD}. See Line.7 in Metric.py: \texttt{num\_class = 37}}, however, a lot of previous works incorrectly compute SeK over 7 land cover classes due to historical problems.
Actually, the SECOND dataset only consists of 30 change types, and the kappa-based metric makes it difficult to analyze false positives and false negatives.
To address these issues, we adopt an IoU-based metric: mIoU$_{30}$, which computes the average IoU across all 30 change types.
mIoU$_{30}$ is also class-sensitive, like SeK, but it provides an easier analysis of errors for each change type.

\mpara{1mm}{Result}.
Considering the change detection reality check problem raised by \cite{corley2024change}, we also benchmark a Siamese-UNet as a strong baseline.
Table~\ref{tab:bench_scd} presents a comprehensive benchmark on SECOND for BCD and SCD.
Our results show those contemporary semantic change detection architectures tailored for the nature of change (e.g., the causal relationship between semantics and change: ChangeMask and Bi-SRNet; temporal symmetry: ChangeMask) outperform Siamese-UNet on SCD.
ChangeStar (1$\times$96) with EfficientNet-B0 has comparable performance with Siamese-UNet.
Meanwhile, to align with the aforementioned settings, we continue to utilize ChangeStar for subsequent data-centric studies.

We first compare Changen2 pre-training with OpenEarthMap (OEM) pre-training to ablate OEM's impact since \texttt{Changen2-S9-27k} includes OEM dataset. 
The results suggest that Changen2 pre-training consistently outperforms OEM pre-training across various backbones, including EfficientNet, SwinV2-B, and ViT-B.
This confirms the information gain provided by Changen2.
Compared to Satlas \cite{satlas} and SA-1B \cite{sam} pre-training, Changen2 pre-training still brings superior improvements in both BCD and SCD.
Besides, we have a similar observation that the larger model benefits more from Changen2 pre-training.
For example, using OEM pre-training as a reference, the Efficient-B0 variant, with 9.4M parameters, has a higher SeK but a lower mIoU$_{30}$ compared to its OEM counterpart. However, when using larger backbones like SwinV2-B and ViT-B, both SeK and mIoU$_{30}$ are higher than their OEM counterparts.
Additionally, our zero-shot ChangeStar (1$\times$256) with ViT-B achieves 61.0\% F$_1$, reducing the gap with fully supervised Siamese-UNet to only 9.4\%.
These results confirm the transferability of Changen2 pre-training and the feasibility of our synthetic data pre-training roadmap for zero-shot change detection.

\subsection{Comparison with Previous RS Foundation Models}\label{sec:rsfm}
Existing remote sensing foundation models are mainly based on large-scale supervised pre-training \cite{satlas}, self-supervised contrastive learning \cite{moco, mocov2}, 
self-supervised masked autoencoders (MAE) \cite{mae}, or their combinations.
Our self-supervised Changen2 presents a new roadmap for producing task-tailored RS foundation models, i.e., synthetic data pre-training (SDP).

\begin{table*}[t]
\caption{\textbf{Comparison with state-of-the-art remote sensing foundation models}.
Note that \texttt{Change2-S0-1.2M} is generated from unlabeled fMoW satellite images, and the change data generator, Changen2, is trained in a self-supervised manner on these unlabeled images.
This means that our Changen2 uses exactly the same data as those foundation models trained on fMoW.
All change detection architectures are ChangeStar (1$\times$256).
\label{tab:fm_comp}}
\centering
\small
\tablestyle{4pt}{1.4}
\resizebox{\linewidth}{!}{
\begin{tabular}{l|c|cc|ccc|ccc|ccc|cc}
&  &      &                      & \multicolumn{3}{c|}{LEVIR-CD} & \multicolumn{3}{c|}{S2Looking} & \multicolumn{3}{c|}{xView2} & \multicolumn{2}{c}{SECOND} \\
Method & Ref. & Pre-training data  & Backbone  & F$_1$ & Prec. & Rec. & F$_1$ & Prec. & Rec. & F$_1^{\rm all}$ & F$_1^{\rm loc}$& F$_1^{\rm dam}$  & SeK$_{37}$ &  mIoU$_{30}$ \\
\shline
\multicolumn{2}{l|}{\ft\textit{supervised approaches}}&&&&&&&&&&&& \\
Satlas \cite{satlas}&ICCV'23&Satlas-HiRes&SwinV2-B&91.7&92.5&90.9&\underline{68.3}&72.6&64.5&\underline{78.4}&87.7&74.4&\underline{21.8}&28.1 \\
SAM \cite{sam}&ICCV'23&SA-1B&ViT-B&\underline{92.0}&94.2&90.0&68.1&71.5&64.9&\textbf{78.5}&88.1&74.4&21.5&\underline{30.0}\\
\multicolumn{2}{l|}{\zs\textit{self-supervised approaches}}& &  &  & &&&&&&&& \\
GASSL \cite{gassl}&ICCV'21&fMoW&ResNet-50&89.6&90.6&88.6&66.3&71.2&62.1&75.5&86.8&70.8&18.3&25.9\\
SeCo \cite{seco}&ICCV'21&SeCo-1M&ResNet-50 & 88.4 & 89.5&87.4&66.0&70.3&62.2&75.7&86.2&71.2&17.0&23.3\\
SatMAE \cite{satmae}&NeurIPS'22&fMoW&ViT-L&90.0&91.9&88.2&65.0&72.0&59.2&75.9&87.5&70.9&20.1&25.4\\
CACo \cite{caco}&CVPR'23&CACo-1M&ResNet-50&89.2&90.4&88.0&65.9&70.0&62.2&75.5&86.7&70.8&17.0&24.1 \\
Scale-MAE \cite{scalemae}&ICCV'23&fMoW&ViT-L&86.6&89.1&84.3&50.2&72.3&32.5&68.5&82.9&62.3&20.6&26.7\\
Cross-Scale MAE \cite{csmae}&NeurIPS'23&fMoW&ViT-L&88.5&89.4&87.6&61.2&66.9&56.5&73.7&85.5&68.6&16.8&21.9\\
SatMAE++ \cite{satmaepp} &CVPR'24&fMoW&ViT-L&90.7&92.2&89.1&56.4&60.8&52.5&75.2&86.9&70.2&19.3&23.9\\
\gr Changen2 (ours) & - &\texttt{Changen2-S0-1.2M}&ViT-B&\textbf{92.2}&93.1&91.4&\textbf{69.1}&71.0&67.3&78.1&87.9&73.9&\textbf{24.0}&\textbf{31.5}\\
\end{tabular}}
\end{table*}

\mpara{1mm}{Setup.}
To comprehensively evaluate our self-supervised Changen2 pre-training, we benchmark it with two state-of-the-art supervised pre-training visual foundation models: Satlas pre-trained SwinV2-B \cite{satlas} and SA-1B pre-trained on ViT-B \cite{sam}, as well as seven state-of-the-art self-supervised remote sensing foundation models. 
These include three contrastive learning-based models: GASSL \cite{gassl}, SeCo \cite{seco}, and CACo \cite{caco}, and four MAE-based models: SatMAE \cite{satmae}, Scale-MAE \cite{scalemae}, Cross-Scale MAE \cite{csmae}, and SatMAE++ \cite{satmaepp}.
We adapt these foundation models to four different change detection tasks: ordinary building change detection, globally distributed off-nadir building change detection, globally distributed building damage assessment (one-to-many semantic change detection), and urban land use/land cover change detection (many-to-many semantic change detection). 
This is done through fully supervised fine-tuning, following common practice \cite{caco, satlas, scalemae}. 
For a fair comparison, all change detection architectures are ChangeStar (1$\times$256) using these foundation models as the backbone.
The training details follow Sec.~\ref{sec:sdp} for LEVIR-CD, S2Looking, and SECOND.
For xView2 dataset, we fine-tune each model with 80k steps, and other settings are the same as in Sec.~\ref{sec:sdp}.
The fine-tuning settings are identical for each model.
The only difference lies in the foundation models themselves, ensuring rigorous benchmarking.

\mpara{1mm}{Self-supervised change data synthesis}.
We leverage a self-supervised Changen2 to generate change data (bitemporal image pair and its dense change label, as shown in Fig.~\ref{fig:ssl_case}) only from unlabeled single-temporal images via Eq.~\ref{eq:ssl_g}.
The generated change training samples are class-agnostic.
By default, we use unlabeled images of fMoW \cite{fmow}, which comprises globally distributed high spatial resolution satellite images, as our single-temporal images to create self-supervised synthetic change data.
After applying sliding window cropping to fMoW, we obtain 1,278,185 images with 512$\times$512 px.
Using these images, we generate a class-agnostic change detection dataset for synthetic data pre-training, referred to as \texttt{Changen2-S0-1.2M}.

\mpara{1mm}{Self-supervised Changen2 pre-training details}.
Using \texttt{Change2-S0-1.2M},
we integrally pre-train ChangeStar (1$\times$256, ViT-B) for 800k steps.
Other pre-training settings are the same as Sec.~\ref{sec:sdp}.
The self-supervised Changen2 is trained using unlabeled 1.2 million fMoW satellite images. 
This means that our Changen2 uses exactly the same data as those foundation models trained on fMoW.

\mpara{1mm}{Result}.
Table~\ref{tab:fm_comp} presents a comprehensive benchmark for these foundation models on four different change detection tasks.
Among self-supervised approaches, Changen2 pre-trained foundation models significantly outperform MAE-based and contrastive learning-based remote sensing foundation models by a large margin. 
For example, on LEVIR-CD, our model achieves 92.2\% F$_1$ compared to the previous best of 90.7\% (SatMAE++); on S2Looking, our model scores 69.1\% F$_1$ versus the previous best of 66.3\% (GASSL); on xView2, our model reaches 78.1\% F$_1^{\rm all}$ compared to the previous best of 75.9\% (SatMAE); and on SECOND, our model achieves 24.0\% SeK$_{37}$ and 31.5\% mIoU$_{30}$ versus the previous best of 20.6\% SeK$_{37}$ and 26.7\% mIoU$_{30}$ (Scale-MAE).
Compared to the two supervised foundation models, self-supervised Changen2 pre-trained foundation models still outperform SAM and Satlas on LEVIR-CD, S2Looking, and SECOND, and demonstrates comparable performance on xView2.

\mpara{1mm}{Discussion}.
From results of Table~\ref{tab:fm_comp}, we have the following important observations:
(i) the above seven self-supervised RSFMs consistently lag behind the supervised visual FMs in change detection performances.
Our self-supervised Changen2 pre-trained FM, as a change-tailored RSFM, bridges the performance gap with supervised visual FMs, for the first time.
This result indicates that developing task-tailored foundation models is worthwhile, at least in the Earth observation domain.
(ii) Self-supervised Changen2 pre-training can outperform supervised Changen2 pre-training on S2Looking (F$_1$: 69.1\% vs. 68.5\%) and SECOND benchmarks (SeK$_{37}$: 24.0 vs. 23.0, mIoU$_{30}$: 31.5 vs. 30.4).
This observation highlights the importance of global image distribution and large data volume to RSFMs and Earth observation applications.

\section{Conclusion}
\label{sec:conc}

We propose a novel and practical data-centric roadmap called generative change modeling to address the ``data hunger'' problem in multi-temporal remote sensing. 
To implement this approach, we introduce Changen2, a generative change foundation model that can generate time series of remote sensing images along with corresponding semantic and change labels from both labeled and unlabeled single-temporal images. 
Changen2 can be trained at scale using our self-supervised learning method, which effectively derives change supervisory signals from unlabeled single images. 
Orthogonal to existing ``foundation models'', our generative change foundation model creates task-tailored foundation models specifically for change detection, equipped with inherent zero-shot change detection capabilities and excellent transferability.

{\small
   \bibliographystyle{IEEEtran}
   \bibliography{main}

\begin{thebibliography}{10}
\providecommand{\url}[1]{#1}
\csname url@samestyle\endcsname
\providecommand{\newblock}{\relax}
\providecommand{\bibinfo}[2]{#2}
\providecommand{\BIBentrySTDinterwordspacing}{\spaceskip=0pt\relax}
\providecommand{\BIBentryALTinterwordstretchfactor}{4}
\providecommand{\BIBentryALTinterwordspacing}{\spaceskip=\fontdimen2\font plus
\BIBentryALTinterwordstretchfactor\fontdimen3\font minus \fontdimen4\font\relax}
\providecommand{\BIBforeignlanguage}[2]{{%
\expandafter\ifx\csname l@#1\endcsname\relax
\typeout{** WARNING: IEEEtran.bst: No hyphenation pattern has been}%
\typeout{** loaded for the language `#1'. Using the pattern for}%
\typeout{** the default language instead.}%
\else
\language=\csname l@#1\endcsname
\fi
#2}}
\providecommand{\BIBdecl}{\relax}
\BIBdecl

\bibitem{changen}
Z.~Zheng, S.~Tian, A.~Ma, L.~Zhang, and Y.~Zhong, ``Scalable multi-temporal remote sensing change data generation via simulating stochastic change process,'' in \emph{ICCV}, 2023, pp. 21\,818--21\,827.

\bibitem{daudt2018fully}
R.~C. Daudt, B.~Le~Saux, and A.~Boulch, ``Fully convolutional siamese networks for change detection,'' in \emph{ICIP}.\hskip 1em plus 0.5em minus 0.4em\relax IEEE, 2018, pp. 4063--4067.

\bibitem{changestar}
Z.~Zheng, A.~Ma, L.~Zhang, and Y.~Zhong, ``Change is {E}verywhere: Single-temporal supervised object change detection in remote sensing imagery,'' in \emph{ICCV}, 2021, pp. 15\,193--15\,202.

\bibitem{changeos}
Z.~Zheng, Y.~Zhong, J.~Wang, A.~Ma, and L.~Zhang, ``Building damage assessment for rapid disaster response with a deep object-based semantic change detection framework: From natural disasters to man-made disasters,'' \emph{Remote Sensing of Environment}, vol. 265, p. 112636, 2021.

\bibitem{zheng2022changemask}
Z.~Zheng, Y.~Zhong, S.~Tian, A.~Ma, and L.~Zhang, ``Change{M}ask: Deep multi-task encoder-transformer-decoder architecture for semantic change detection,'' \emph{ISPRS Journal of Photogrammetry and Remote Sensing}, vol. 183, pp. 228--239, 2022.

\bibitem{siamese}
J.~Bromley, I.~Guyon, Y.~LeCun, E.~S{\"a}ckinger, and R.~Shah, ``Signature verification using a" siamese" time delay neural network,'' in \emph{NeurIPS}, 1994.

\bibitem{gupta2019xbd}
R.~Gupta, R.~Hosfelt, S.~Sajeev, N.~Patel, B.~Goodman, J.~Doshi, E.~Heim, H.~Choset, and M.~Gaston, ``xbd: A dataset for assessing building damage from satellite imagery,'' \emph{arXiv preprint arXiv:1911.09296}, 2019.

\bibitem{shen2021s2looking}
L.~Shen, Y.~Lu, H.~Chen, H.~Wei, D.~Xie, J.~Yue, R.~Chen, S.~Lv, and B.~Jiang, ``S2looking: A satellite side-looking dataset for building change detection,'' \emph{Remote Sensing}, vol.~13, no.~24, p. 5094, 2021.

\bibitem{DEN}
A.~Toker, L.~Kondmann, M.~Weber, M.~Eisenberger, A.~Camero, J.~Hu, A.~P. Hoderlein, {\c{C}}.~{\c{S}}enaras, T.~Davis, D.~Cremers \emph{et~al.}, ``Dynamic{E}arth{N}et: Daily multi-spectral satellite dataset for semantic change segmentation,'' in \emph{CVPR}, 2022, pp. 21\,158--21\,167.

\bibitem{tian2022large}
S.~Tian, Y.~Zhong, Z.~Zheng, A.~Ma, X.~Tan, and L.~Zhang, ``Large-scale deep learning based binary and semantic change detection in ultra high resolution remote sensing imagery: From benchmark datasets to urban application,'' \emph{ISPRS Journal of Photogrammetry and Remote Sensing}, vol. 193, pp. 164--186, 2022.

\bibitem{bourdis2011constrained}
N.~Bourdis, D.~Marraud, and H.~Sahbi, ``Constrained optical flow for aerial image change detection,'' in \emph{2011 IEEE International Geoscience and Remote Sensing Symposium}.\hskip 1em plus 0.5em minus 0.4em\relax IEEE, 2011, pp. 4176--4179.

\bibitem{kolos2019procedural}
M.~Kolos, A.~Marin, A.~Artemov, and E.~Burnaev, ``Procedural synthesis of remote sensing images for robust change detection with neural networks,'' in \emph{International Symposium on Neural Networks}.\hskip 1em plus 0.5em minus 0.4em\relax Springer, 2019, pp. 371--387.

\bibitem{song2024syntheworld}
J.~Song, H.~Chen, and N.~Yokoya, ``Syntheworld: A large-scale synthetic dataset for land cover mapping and building change detection,'' 2024, pp. 8287--8296.

\bibitem{copypaste}
G.~Ghiasi, Y.~Cui, A.~Srinivas, R.~Qian, T.-Y. Lin, E.~D. Cubuk, Q.~V. Le, and B.~Zoph, ``Simple copy-paste is a strong data augmentation method for instance segmentation,'' in \emph{CVPR}, 2021, pp. 2918--2928.

\bibitem{chen2021adversarial}
H.~Chen, W.~Li, and Z.~Shi, ``Adversarial instance augmentation for building change detection in remote sensing images,'' \emph{IEEE Transactions on Geoscience and Remote Sensing}, vol.~60, pp. 1--16, 2021.

\bibitem{sam}
A.~Kirillov, E.~Mintun, N.~Ravi, H.~Mao, C.~Rolland, L.~Gustafson, T.~Xiao, S.~Whitehead, A.~C. Berg, W.-Y. Lo, P.~Dollar, and R.~Girshick, ``Segment anything,'' in \emph{ICCV}, October 2023, pp. 4015--4026.

\bibitem{diffusionsat}
S.~Khanna, P.~Liu, L.~Zhou, C.~Meng, R.~Rombach, M.~Burke, D.~B. Lobell, and S.~Ermon, ``Diffusionsat: A generative foundation model for satellite imagery,'' in \emph{ICLR}, 2024.

\bibitem{satmae}
Y.~Cong, S.~Khanna, C.~Meng, P.~Liu, E.~Rozi, Y.~He, M.~Burke, D.~Lobell, and S.~Ermon, ``Satmae: Pre-training transformers for temporal and multi-spectral satellite imagery,'' vol.~35, 2022, pp. 197--211.

\bibitem{satlas}
F.~Bastani, P.~Wolters, R.~Gupta, J.~Ferdinando, and A.~Kembhavi, ``Satlaspretrain: A large-scale dataset for remote sensing image understanding,'' in \emph{ICCV}, 2023, pp. 16\,772--16\,782.

\bibitem{anychange}
Z.~Zheng, Y.~Zhong, L.~Zhang, and S.~Ermon, ``Segment any change,'' \emph{arXiv preprint arXiv:2402.01188}, 2024.

\bibitem{achiam2023gpt}
J.~Achiam, S.~Adler, S.~Agarwal, L.~Ahmad, I.~Akkaya, F.~L. Aleman, D.~Almeida, J.~Altenschmidt, S.~Altman, S.~Anadkat \emph{et~al.}, ``Gpt-4 technical report,'' \emph{arXiv preprint arXiv:2303.08774}, 2023.

\bibitem{rombach2022high}
R.~Rombach, A.~Blattmann, D.~Lorenz, P.~Esser, and B.~Ommer, ``High-resolution image synthesis with latent diffusion models,'' in \emph{Proceedings of the IEEE/CVF conference on computer vision and pattern recognition}, 2022, pp. 10\,684--10\,695.

\bibitem{spade}
T.~Park, M.-Y. Liu, T.-C. Wang, and J.-Y. Zhu, ``Semantic image synthesis with spatially-adaptive normalization,'' in \emph{CVPR}, 2019, pp. 2337--2346.

\bibitem{pix2pixHD}
T.-C. Wang, M.-Y. Liu, J.-Y. Zhu, A.~Tao, J.~Kautz, and B.~Catanzaro, ``High-resolution image synthesis and semantic manipulation with conditional gans,'' in \emph{CVPR}, 2018, pp. 8798--8807.

\bibitem{zhu2020sean}
P.~Zhu, R.~Abdal, Y.~Qin, and P.~Wonka, ``{SEAN}: Image synthesis with semantic region-adaptive normalization,'' in \emph{CVPR}, 2020, pp. 5104--5113.

\bibitem{tan2021diverse}
Z.~Tan, M.~Chai, D.~Chen, J.~Liao, Q.~Chu, B.~Liu, G.~Hua, and N.~Yu, ``Diverse semantic image synthesis via probability distribution modeling,'' in \emph{CVPR}, 2021, pp. 7962--7971.

\bibitem{shi2022retrieval}
Y.~Shi, X.~Liu, Y.~Wei, Z.~Wu, and W.~Zuo, ``Retrieval-based spatially adaptive normalization for semantic image synthesis,'' in \emph{CVPR}, 2022, pp. 11\,224--11\,233.

\bibitem{oasis}
E.~Sch{\"o}nfeld, V.~Sushko, D.~Zhang, J.~Gall, B.~Schiele, and A.~Khoreva, ``You only need adversarial supervision for semantic image synthesis,'' in \emph{ICLR}, 2021.

\bibitem{cGAN}
M.~Mirza and S.~Osindero, ``Conditional generative adversarial nets,'' \emph{arXiv preprint arXiv:1411.1784}, 2014.

\bibitem{pix2pix}
P.~Isola, J.-Y. Zhu, T.~Zhou, and A.~A. Efros, ``Image-to-image translation with conditional adversarial networks,'' in \emph{CVPR}, 2017, pp. 1125--1134.

\bibitem{dosovitskiy2016generating}
A.~Dosovitskiy and T.~Brox, ``Generating images with perceptual similarity metrics based on deep networks,'' in \emph{NeurIPS}, vol.~29, 2016.

\bibitem{johnson2016perceptual}
J.~Johnson, A.~Alahi, and L.~Fei-Fei, ``Perceptual losses for real-time style transfer and super-resolution,'' in \emph{ECCV}.\hskip 1em plus 0.5em minus 0.4em\relax Springer, 2016, pp. 694--711.

\bibitem{ddpm}
J.~Ho, A.~Jain, and P.~Abbeel, ``Denoising diffusion probabilistic models,'' \emph{NeurIPS}, vol.~33, pp. 6840--6851, 2020.

\bibitem{sohl2015deep}
J.~Sohl-Dickstein, E.~Weiss, N.~Maheswaranathan, and S.~Ganguli, ``Deep unsupervised learning using nonequilibrium thermodynamics,'' in \emph{International conference on machine learning}.\hskip 1em plus 0.5em minus 0.4em\relax PMLR, 2015, pp. 2256--2265.

\bibitem{hyvarinen2005estimation}
A.~Hyv{\"a}rinen and P.~Dayan, ``Estimation of non-normalized statistical models by score matching.'' \emph{Journal of Machine Learning Research}, vol.~6, no.~4, 2005.

\bibitem{song2019generative}
Y.~Song and S.~Ermon, ``Generative modeling by estimating gradients of the data distribution,'' \emph{Advances in neural information processing systems}, vol.~32, 2019.

\bibitem{song2021scorebased}
Y.~Song, J.~Sohl-Dickstein, D.~P. Kingma, A.~Kumar, S.~Ermon, and B.~Poole, ``Score-based generative modeling through stochastic differential equations,'' in \emph{ICLR}, 2021.

\bibitem{controlnet}
L.~Zhang, A.~Rao, and M.~Agrawala, ``Adding conditional control to text-to-image diffusion models,'' in \emph{ICCV}, 2023, pp. 3836--3847.

\bibitem{uvit}
F.~Bao, S.~Nie, K.~Xue, Y.~Cao, C.~Li, H.~Su, and J.~Zhu, ``All are worth words: A vit backbone for diffusion models,'' in \emph{Proceedings of the IEEE/CVF Conference on Computer Vision and Pattern Recognition}, 2023, pp. 22\,669--22\,679.

\bibitem{dit}
W.~Peebles and S.~Xie, ``Scalable diffusion models with transformers,'' in \emph{ICCV}, 2023, pp. 4195--4205.

\bibitem{moco}
K.~He, H.~Fan, Y.~Wu, S.~Xie, and R.~Girshick, ``Momentum contrast for unsupervised visual representation learning,'' in \emph{CVPR}, 2020, pp. 9729--9738.

\bibitem{mocov2}
X.~Chen, H.~Fan, R.~Girshick, and K.~He, ``Improved baselines with momentum contrastive learning,'' \emph{arXiv preprint arXiv:2003.04297}, 2020.

\bibitem{beit}
H.~Bao, L.~Dong, S.~Piao, and F.~Wei, ``{BE}it: {BERT} pre-training of image transformers,'' in \emph{International Conference on Learning Representations}, 2022.

\bibitem{simmim}
Z.~Xie, Z.~Zhang, Y.~Cao, Y.~Lin, J.~Bao, Z.~Yao, Q.~Dai, and H.~Hu, ``Simmim: A simple framework for masked image modeling,'' in \emph{CVPR}, 2022, pp. 9653--9663.

\bibitem{mae}
K.~He, X.~Chen, S.~Xie, Y.~Li, P.~Doll{\'a}r, and R.~Girshick, ``Masked autoencoders are scalable vision learners,'' in \emph{CVPR}, 2022, pp. 16\,000--16\,009.

\bibitem{gassl}
K.~Ayush, B.~Uzkent, C.~Meng, K.~Tanmay, M.~Burke, D.~Lobell, and S.~Ermon, ``Geography-aware self-supervised learning,'' in \emph{ICCV}, 2021, pp. 10\,181--10\,190.

\bibitem{seco}
O.~Manas, A.~Lacoste, X.~Gir{\'o}-i Nieto, D.~Vazquez, and P.~Rodriguez, ``Seasonal contrast: Unsupervised pre-training from uncurated remote sensing data,'' in \emph{ICCV}, 2021, pp. 9414--9423.

\bibitem{caco}
U.~Mall, B.~Hariharan, and K.~Bala, ``Change-aware sampling and contrastive learning for satellite images,'' in \emph{CVPR}, 2023, pp. 5261--5270.

\bibitem{hong2024spectralgpt}
D.~Hong, B.~Zhang, X.~Li, Y.~Li, C.~Li, J.~Yao, N.~Yokoya, H.~Li, P.~Ghamisi, X.~Jia \emph{et~al.}, ``Spectralgpt: Spectral remote sensing foundation model,'' \emph{IEEE TPAMI}, 2024.

\bibitem{feichtenhofer2022masked}
C.~Feichtenhofer, Y.~Li, K.~He \emph{et~al.}, ``Masked autoencoders as spatiotemporal learners,'' \emph{Advances in neural information processing systems}, vol.~35, pp. 35\,946--35\,958, 2022.

\bibitem{tong2022videomae}
Z.~Tong, Y.~Song, J.~Wang, and L.~Wang, ``Videomae: Masked autoencoders are data-efficient learners for self-supervised video pre-training,'' \emph{Advances in neural information processing systems}, vol.~35, pp. 10\,078--10\,093, 2022.

\bibitem{scalemae}
C.~J. Reed, R.~Gupta, S.~Li, S.~Brockman, C.~Funk, B.~Clipp, K.~Keutzer, S.~Candido, M.~Uyttendaele, and T.~Darrell, ``Scale-mae: A scale-aware masked autoencoder for multiscale geospatial representation learning,'' in \emph{ICCV}, 2023, pp. 4088--4099.

\bibitem{csmae}
M.~Tang, A.~Cozma, K.~Georgiou, and H.~Qi, ``Cross-scale mae: A tale of multiscale exploitation in remote sensing,'' \emph{Advances in Neural Information Processing Systems}, vol.~36, 2024.

\bibitem{satmaepp}
M.~Noman, M.~Naseer, H.~Cholakkal, R.~M. Anwar, S.~Khan, and F.~S. Khan, ``Rethinking transformers pre-training for multi-spectral satellite imagery,'' \emph{arXiv preprint arXiv:2403.05419}, 2024.

\bibitem{clip}
A.~Radford, J.~W. Kim, C.~Hallacy, A.~Ramesh, G.~Goh, S.~Agarwal, G.~Sastry, A.~Askell, P.~Mishkin, J.~Clark \emph{et~al.}, ``Learning transferable visual models from natural language supervision,'' in \emph{International conference on machine learning}.\hskip 1em plus 0.5em minus 0.4em\relax PMLR, 2021, pp. 8748--8763.

\bibitem{gpt2}
A.~Radford, J.~Wu, R.~Child, D.~Luan, D.~Amodei, I.~Sutskever \emph{et~al.}, ``Language models are unsupervised multitask learners,'' \emph{OpenAI blog}, vol.~1, no.~8, p.~9, 2019.

\bibitem{gpt4}
J.~Achiam, S.~Adler, S.~Agarwal, L.~Ahmad, I.~Akkaya, F.~L. Aleman, D.~Almeida, J.~Altenschmidt, S.~Altman, S.~Anadkat \emph{et~al.}, ``Gpt-4 technical report,'' \emph{arXiv preprint arXiv:2303.08774}, 2023.

\bibitem{oem}
J.~Xia, N.~Yokoya, B.~Adriano, and C.~Broni-Bediako, ``Openearthmap: A benchmark dataset for global high-resolution land cover mapping,'' in \emph{Proceedings of the IEEE/CVF Winter Conference on Applications of Computer Vision}, 2023, pp. 6254--6264.

\bibitem{multidiffusion}
O.~Bar-Tal, L.~Yariv, Y.~Lipman, and T.~Dekel, ``Multidiffusion: Fusing diffusion paths for controlled image generation,'' \emph{ICML}, 2023.

\bibitem{kim2024pagoda}
D.~Kim, C.-H. Lai, W.-H. Liao, Y.~Takida, N.~Murata, T.~Uesaka, Y.~Mitsufuji, and S.~Ermon, ``Pagoda: Progressive growing of a one-step generator from a low-resolution diffusion teacher,'' \emph{arXiv preprint arXiv:2405.14822}, 2024.

\bibitem{islam2020much}
M.~A. Islam, S.~Jia, and N.~D. Bruce, ``How much position information do convolutional neural networks encode?'' \emph{arXiv preprint arXiv:2001.08248}, 2020.

\bibitem{kingma2013auto}
D.~P. Kingma and M.~Welling, ``Auto-encoding variational bayes,'' \emph{arXiv preprint arXiv:1312.6114}, 2013.

\bibitem{nichol2021improved}
A.~Q. Nichol and P.~Dhariwal, ``Improved denoising diffusion probabilistic models,'' in \emph{International conference on machine learning}.\hskip 1em plus 0.5em minus 0.4em\relax PMLR, 2021, pp. 8162--8171.

\bibitem{lugmayr2022repaint}
A.~Lugmayr, M.~Danelljan, A.~Romero, F.~Yu, R.~Timofte, and L.~Van~Gool, ``Repaint: Inpainting using denoising diffusion probabilistic models,'' in \emph{Proceedings of the IEEE/CVF conference on computer vision and pattern recognition}, 2022, pp. 11\,461--11\,471.

\bibitem{ddim}
J.~Song, C.~Meng, and S.~Ermon, ``Denoising diffusion implicit models,'' in \emph{International Conference on Learning Representations}, 2021.

\bibitem{FID}
M.~Heusel, H.~Ramsauer, T.~Unterthiner, B.~Nessler, and S.~Hochreiter, ``Gans trained by a two time-scale update rule converge to a local nash equilibrium,'' in \emph{NeurIPS}, vol.~30, 2017.

\bibitem{IS}
T.~Salimans, I.~Goodfellow, W.~Zaremba, V.~Cheung, A.~Radford, and X.~Chen, ``Improved techniques for training gans,'' in \emph{NeurIPS}, vol.~29, 2016.

\bibitem{loshchilov2017decoupled}
I.~Loshchilov and F.~Hutter, ``Decoupled weight decay regularization,'' \emph{arXiv preprint arXiv:1711.05101}, 2017.

\bibitem{zheng2023farseg++}
Z.~Zheng, Y.~Zhong, J.~Wang, A.~Ma, and L.~Zhang, ``Far{S}eg++: Foreground-aware relation network for geospatial object segmentation in high spatial resolution remote sensing imagery,'' \emph{IEEE TPAMI}, 2023.

\bibitem{levircd}
H.~Chen and Z.~Shi, ``A spatial-temporal attention-based method and a new dataset for remote sensing image change detection,'' \emph{Remote Sensing}, vol.~12, no.~10, p. 1662, 2020.

\bibitem{whucd}
S.~Ji, S.~Wei, and M.~Lu, ``Fully convolutional networks for multisource building extraction from an open aerial and satellite imagery data set,'' \emph{IEEE Transactions on Geoscience and Remote Sensing}, vol.~57, no.~1, pp. 574--586, 2018.

\bibitem{ade20k}
B.~Zhou, H.~Zhao, X.~Puig, T.~Xiao, S.~Fidler, A.~Barriuso, and A.~Torralba, ``Semantic understanding of scenes through the ade20k dataset,'' \emph{IJCV}, vol. 127, no.~3, pp. 302--321, 2019.

\bibitem{changeformer}
W.~G.~C. Bandara and V.~M. Patel, ``A transformer-based siamese network for change detection,'' in \emph{IGARSS}, 2022, pp. 207--210.

\bibitem{chen2021remote}
H.~Chen, Z.~Qi, and Z.~Shi, ``Remote sensing image change detection with transformers,'' \emph{IEEE Transactions on Geoscience and Remote Sensing}, vol.~60, pp. 1--14, 2021.

\bibitem{wang2022empirical}
D.~Wang, J.~Zhang, B.~Du, G.-S. Xia, and D.~Tao, ``An empirical study of remote sensing pretraining,'' \emph{IEEE Transactions on Geoscience and Remote Sensing}, 2022.

\bibitem{li2023new}
K.~Li, X.~Cao, and D.~Meng, ``A new learning paradigm for foundation model-based remote sensing change detection,'' \emph{arXiv preprint arXiv:2312.01163}, 2023.

\bibitem{rdpnet}
H.~Chen, F.~Pu, R.~Yang, R.~Tang, and X.~Xu, ``Rdp-net: Region detail preserving network for change detection,'' \emph{IEEE Transactions on Geoscience and Remote Sensing}, vol.~60, pp. 1--10, 2022.

\bibitem{a2net}
Z.~Li, C.~Tang, X.~Liu, W.~Zhang, J.~Dou, L.~Wang, and A.~Y. Zomaya, ``Lightweight remote sensing change detection with progressive feature aggregation and supervised attention,'' \emph{IEEE Transactions on Geoscience and Remote Sensing}, vol.~61, pp. 1--12, 2023.

\bibitem{ding2022bi}
L.~Ding, H.~Guo, S.~Liu, L.~Mou, J.~Zhang, and L.~Bruzzone, ``Bi-temporal semantic reasoning for the semantic change detection in hr remote sensing images,'' \emph{IEEE Transactions on Geoscience and Remote Sensing}, vol.~60, pp. 1--14, 2022.

\bibitem{dinov2}
M.~Oquab, T.~Darcet, T.~Moutakanni, H.~V. Vo, M.~Szafraniec, V.~Khalidov, P.~Fernandez, D.~Haziza, F.~Massa, A.~El-Nouby, R.~Howes, P.-Y. Huang, H.~Xu, V.~Sharma, S.-W. Li, W.~Galuba, M.~Rabbat, M.~Assran, N.~Ballas, G.~Synnaeve, I.~Misra, H.~Jegou, J.~Mairal, P.~Labatut, A.~Joulin, and P.~Bojanowski, ``Dinov2: Learning robust visual features without supervision,'' 2023.

\bibitem{second}
K.~Yang, G.-S. Xia, Z.~Liu, B.~Du, W.~Yang, M.~Pelillo, and L.~Zhang, ``Asymmetric siamese networks for semantic change detection in aerial images,'' \emph{IEEE Transactions on Geoscience and Remote Sensing}, vol.~60, pp. 1--18, 2021.

\bibitem{corley2024change}
I.~Corley, C.~Robinson, and A.~Ortiz, ``A change detection reality check,'' \emph{arXiv preprint arXiv:2402.06994}, 2024.

\bibitem{fmow}
G.~Christie, N.~Fendley, J.~Wilson, and R.~Mukherjee, ``Functional map of the world,'' in \emph{Proceedings of the IEEE Conference on Computer Vision and Pattern Recognition}, 2018, pp. 6172--6180.

\end{thebibliography}
}

\end{document}